%% file: template.tex
\documentclass{article}

\usepackage{arxiv}

\usepackage[utf8]{inputenc} 
\usepackage[T1]{fontenc}    
\usepackage{booktabs}       
\usepackage{amsfonts}       
\usepackage{nicefrac}       
\usepackage{microtype}      

\usepackage{amssymb}
\usepackage{latexsym}

\usepackage{url}

\usepackage{xcolor}
\usepackage{hyperref}

\usepackage{graphicx}
\usepackage{tikz}
\usepackage{environ}
\usepackage{layouts}
\usepackage{natbib}
\title{Harnessing spatial homogeneity of neuroimaging data: patch individual filter layers for CNNs}

\author{
    Fabian Eitel \\
    Charit\'e -- Universit\"atsmedizin Berlin\\
    10117 Berlin, Germany\\
   \And
    Jan Philipp Albrecht \\
    Charit\'e -- Universit\"atsmedizin Berlin\\
    10117 Berlin, Germany\\
   \And
    Martin Weygandt \\
    Charit\'e -- Universit\"atsmedizin Berlin\\
    10117 Berlin, Germany\\
   \And
    Friedemann Paul \\
    Charit\'e -- Universit\"atsmedizin Berlin\\
    10117 Berlin, Germany\\
   \And
    Kerstin Ritter \\
    Charit\'e -- Universit\"atsmedizin Berlin\\
    10117 Berlin, Germany\\
}

\begin{document}
\maketitle

\begin{abstract}
Neuroimaging data, e.g. obtained from magnetic resonance imaging (MRI), is comparably homogeneous due to (1) the uniform structure of the brain and (2) additional efforts to spatially normalize the data to a standard template using linear and non-linear transformations. 
Convolutional neural networks (CNNs), in contrast, have been specifically designed for highly heterogeneous data, such as natural images, by sliding convolutional filters over different positions in an image. 
Here, we suggest a new CNN architecture that combines the idea of hierarchical abstraction in neural networks with a prior on the spatial homogeneity of neuroimaging data:
Whereas early layers are trained globally using standard convolutional layers, we introduce for higher, more abstract layers patch individual filters (PIF). By learning filters in individual image regions (patches) without sharing weights, PIF layers can learn abstract features faster and with fewer samples.
We thoroughly evaluated PIF layers for three different tasks and data sets, namely sex classification on UK Biobank data, Alzheimer's disease detection on ADNI data and multiple sclerosis detection on  private hospital data. 
We demonstrate that CNNs using PIF layers result in higher accuracies, especially in low sample size settings, and need fewer training epochs for convergence. 
To the best of our knowledge, this is the first study which introduces a prior on brain MRI for CNN learning.
\end{abstract}

\keywords{convolutional neural networks (CNNs) \and patch individual filters \and neuroimaging \and magnetic resonance imaging (MRI) \and small sample size \and Alzheimer's disease \and multiple sclerosis \and sex classification}

\section{Introduction}
\label{sec:intro}

In recent years, deep learning architectures relying on convolutional neural networks (CNNs) have advanced to a key technology for analyzing medical imaging data from various image sources including magnetic resonance imaging (MRI, e.g. \citet{Litjens2017,VIEIRA2017Review, COLE2017115, LUNDERVOLD2019102}). In neuroimaging, state-of-the-art results have been achieved for diverse pixel-wise segmentation tasks (e.g., segmentation of white matter lesions, brain tumors or vessels; \citet{Kamnitsas2016, KAMNITSAS2017lesions, Livne2019, nair2020exploring}) and image- or volume-wise classification of neurological or psychiatric diseases such as Alzheimer’s disease, multiple sclerosis or schizophrenia \citep{VIEIRA2017Review, korolev2017residual, Rieke2018, Boehle2019, Eitel2019MS}. The models used in most studies here, are largely influenced by architectures which have been shown to be successful in computer vision tasks on natural images \citep{Litjens2017,Guan2019comprehensive, LUNDERVOLD2019102}.

\begin{figure}
    \centering
    \includegraphics[width=0.75\linewidth]{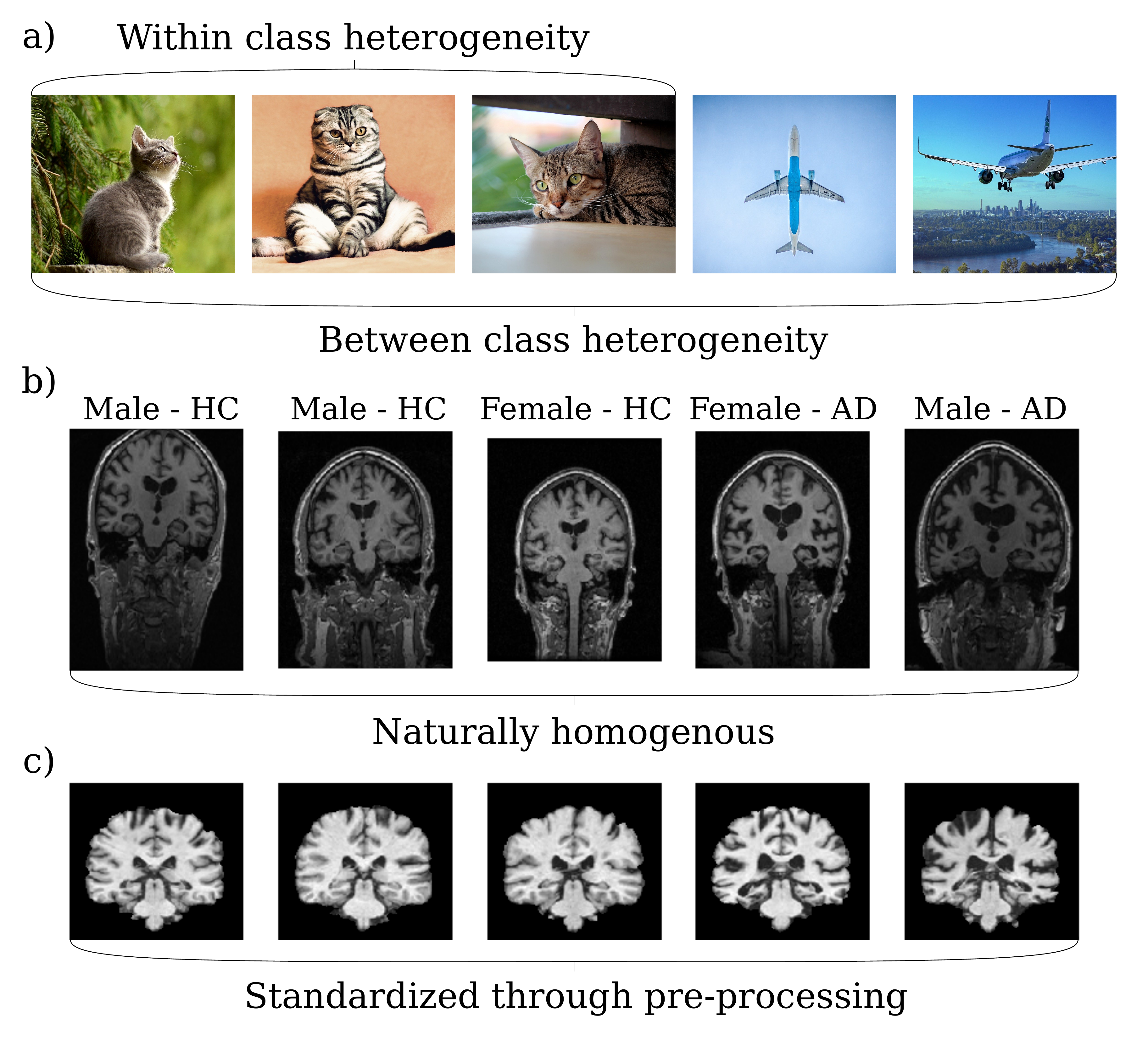}
    \caption{(a) Natural images are typically heterogeneous both within and between classes. (b) MR images of the human brain have homogeneous structures even among different sexes and between healthy subjects (HC) and diseased subjects (AD). (c) Through sophisticated pre-processing techniques, MR images are standardized to a common template reducing their variance further.}
    \label{fig:cats}
\end{figure}

However, in contrast to natural images, neuroimaging data is much more homogeneous (see Figure \ref{fig:cats}) and data sets are typically orders of magnitude smaller. The homogeneity of neuroimaging is due to (1) the inherent structure of the brain, which is mostly identical for individual subjects, i.e. each brain consists of the same parts (cerebellum, frontal lobe, occipital lobe etc.), which are arranged in the same way (e.g., the occipital lobe is in the back). And (2) neuroimaging data is usually further homogenized by normalizing them to a shared template within the MNI space such as the ICBM 152 atlas \citep{john_phd,AVANTS2008,FONOV2011313}. For this, linear and/or non-linear transformations are used and different software packages are available (e.g., SPM\footnote{\url{https://www.fil.ion.ucl.ac.uk/spm}}, FSL\footnote{\url{https://fsl.fmrib.ox.ac.uk/fsl/fslwiki}} or ANTS\footnote{\url{https://github.com/ANTsX/ANTs}}). Generally, this is done to ensure that a voxel at a certain location contains approximately the same brain region in every image and allows researchers to investigate a specific region (e.g., the hippocampus) across subjects. In particular, this is a major prerequisite for mass-univariate as well as multivariate pattern analysis, which have been extensively applied in the neuroimaging domain \citep{Kriegeskorte2006searchlight, Weygandt2011a, Haxby2014mvpa, haynes2015primer}. Acquiring large neuroimaging data sets has strenuous requirements both financially and in terms of expertise, and the strict privacy regulations of medical data in many countries makes the publication of these data sets challenging. Therefore, many machine learning studies are carried out on rather small, local neuroimaging datasets and results often do not generalize \citep{VAROQUAUX2018samplesizes}.

A common method to deal with these small sample sizes is to incorporate known information or assumptions about the data distribution into the learning model. Technically, this can be seen as introducing a prior. In neuroimaging, for instance, studies have incorporated priors into machine learning models by using extracted group-level features, topological structure or other biophysical understandings \citep{Chong2017groupprior, Varoquaux2010NIPS, Woolrich2009bayesian}. However, the application of priors in CNN based neuroimaging studies is not yet common, even though putting highly homogeneous data into standard CNN architectures is sub-optimal. This is due to the fact that computer vision CNNs are optimized to deal with the high spatial variance of natural images (see Figure \ref{fig:cats}a). By using weight-sharing, filters in both early and late layers are being optimized to capture signals regardless of their position. Were all images spatially standardized, i.e. objects are in the same position and have the same angle or viewpoint, it would suffice to search certain abstract objects, such as the ears of a cat, solely within a certain sub-space (i.e., a patch). Although it seems natural to exploit the spatial homogeneity of standardized MR images into a model prior, the technical integration of priors into CNNs is difficult and, to the best of our knowledge, this has not been done yet.

Another method, aimed at improving learning in small sample sizes regimes, is to reduce the amount of features through selection   \citep{Choupan746735}. A model parameter, such as a neural network filter, that is being trained on the entire input will be subject to a greater superimposition of different distributions (from signal and noise) than on a smaller selection of those features, i.e. a sub-space. If we can assume that most selected sub-spaces contain sufficient discriminatory information, then the disentanglement task on each sub-space becomes easier. Therefore, training model parameters on a sub-space of the input should require fewer training samples and iterations for convergence. In Figure \ref{fig:voxels}, we plot typical neuroimaging analysis methods which use non-data-driven feature selection on the spectrum of how many input features each filter or classifier uses. The extreme case of reducing the input would be to fit a model for each voxel individually. This is the case in mass-univariate studies and entails the multiple comparison problem \citep{genovese2002thresholding, POLDRACK2008fMRIGuide, PERNET2015clustercorrection}. A fully-connected neural network is similar in the regard that the weights are learned based on a single input feature and neighbouring information is lost. An intermediate solution would be to train models on regions-of-interest  (ROIs) or, more generally, image patches \citep{greenstein2012ROIschizophrenia, srivastava2019abcd}.

\definecolor{ubuntu_green}{HTML}{88aa00}
\definecolor{ubuntu_blue}{HTML}{5f8dd3}
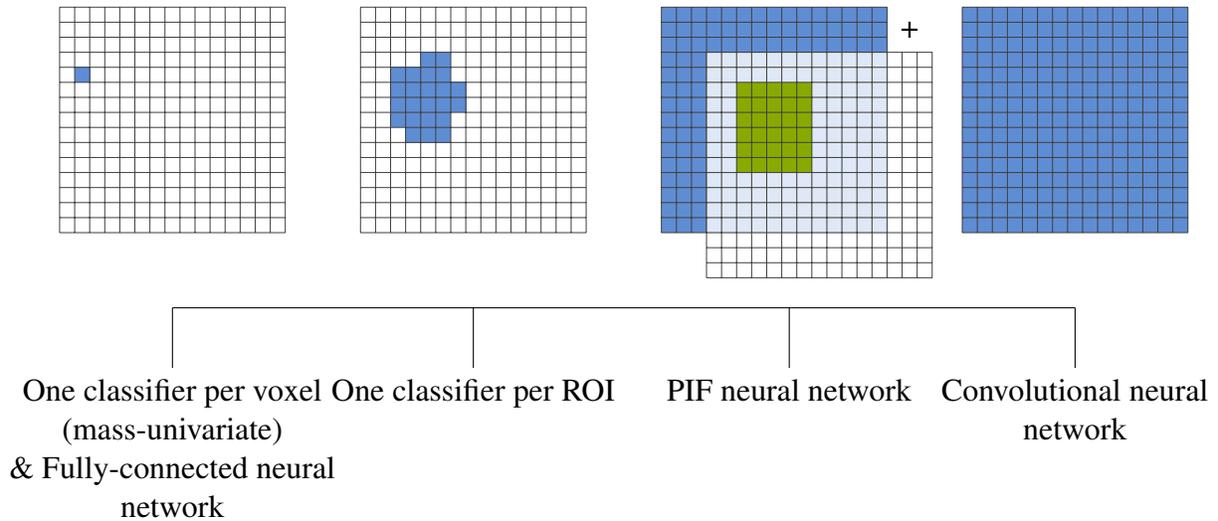
\begin{figure*}
\begin{center}

\begin{tikzpicture}
    \fill[ubuntu_blue] (-2.8,2.0) rectangle (-2.6,2.2);
    \draw[step=0.2cm,darkgray,very thin] (-3,0) grid (0,3);
    \fill[ubuntu_blue] (1.8,2.2) rectangle (2.2,2.4); 
    \fill[ubuntu_blue] (1.4,1.4) rectangle (2.2,2.2); 
    \fill[ubuntu_blue] (1.6,1.2) rectangle (2.2,1.4); 
    \fill[ubuntu_blue] (2.2,1.6) rectangle (2.4,2.0); 
    \draw[step=0.2cm,darkgray,very thin] (0.99999,0) grid (4,3);
    \fill[ubuntu_blue] (5,0) rectangle (8,3);
    \draw[step=0.2cm,darkgray,very thin] (4.999,0) grid (8,3);
    \node at (8.3, 2.7) {+};
    \fill[white, opacity=0.8] (5.6,-0.6) rectangle (8.6,2.4);
    \fill[ubuntu_green] (6.0,0.8) rectangle (7.0,2.0);
    \draw[step=0.2cm,darkgray,very thin] (5.599,-0.6) grid (8.6,2.4);
    \fill[ubuntu_blue] (9,0) rectangle (12,3);
    \draw[step=0.2cm,darkgray,very thin] (8.999,0) grid (12,3);
    \draw (-1.5,-1) -- (10.5,-1);
    \draw (-1.5,-1) -- (-1.5,-1.8) node[anchor=north, align=center] {One classifier per voxel \\
        (mass-univariate) \\
        \& Fully-connected neural \\
        network};
    \draw (2.5,-1) -- (2.5,-1.8) node[anchor=north, align=center] {One classifier per ROI};
    \draw (6.7,-1) -- (6.7,-1.8) node[anchor=north, align=center] {PIF neural network};
    \draw (10.5,-1) -- (10.5,-1.8) node[anchor=north, align=center] {Convolutional neural \\
    network};
\end{tikzpicture}

\caption{Comparison of the number of voxels each feature/kernel uses per model. The grid shows the entire input and in blue/green how much of the input is used in the respective models. Mass-univariate studies use a single voxel per classifier, fully-connected neural networks also use a single weight per voxel albeit combining them after. ROI-based models typically train a single classifier based on an entire ROI or extract a single feature from an ROI. Patch individual filter (PIF) neural networks use both the entire input for lower level features and patches for higher level (latent) features (shown in green). CNN filters use the entire input of each layer throughout the entire network (under some conditions regarding stride and dilation).}
\label{fig:voxels}
\end{center}
\end{figure*}

In this study, we combine a spatial homogeneity prior with feature selection by introducing a new CNN architecture relying on patch individual filter (PIF) layers. In contrast to standard convolutional layers, PIF layers do not perform weight sharing across the entire input but learn individual filters for each location in the data. Since we assume that individual filters are especially relevant for more abstract features, we only applied PIF layers to later layers. For early layers, we used standard convolutional layers to learn globally relevant low-level features such as edges and blobs. Therefore, PIF layers invoke feature selection in the latent space instead of the input space. We evaluated the PIF-architecture with respect to a baseline CNN-architecture for three exemplary tasks within the neuroimaging domain, namely sex classification based on the UK Biobank imaging data\footnote{https://www.ukbiobank.ac.uk/}, Alzheimer's disease (AD) detection based on the Alzheimer's Disease Neuroimaging Initiative (ADNI\footnote{http://adni.loni.usc.edu/}) database and multiple sclerosis (MS) detection based on private data from Charité - Universitätsmedizin Berlin. In most cases, the PIF architecture resulted in a higher balanced accuracy, especially for small data sets. Moreover, because higher level filters in the PIF architecture are trained on patches which contain less information and noise than the entire image, the PIF architecture required much fewer training iterations until convergence.

\section{Methods}
\label{sec:method}

\subsection{Description of PIF layers}
For the analysis of spatially homogeneous and normalized MRI data, we introduce in this section a new CNN architecture relying on PIF layers. Although we perform all experiments in 3D, we describe and visualize here the methods for simplicity in 2D. 
PIF layers consist of 3 stages: (i) split, (ii) process and (iii) reassemble. Each output feature map of the previous layer is first split (i) into patches of size \((s_{x} \times s_{y})\). Next, the patches \(p_{ij}\) centered at row \(i\) and column \(j\) of all feature maps are processed (ii) with a series of local convolutions of kernel size \((k_{x} \times k_{y})\).

\begin{figure}
    \centering
    \includegraphics[width=0.75\linewidth]{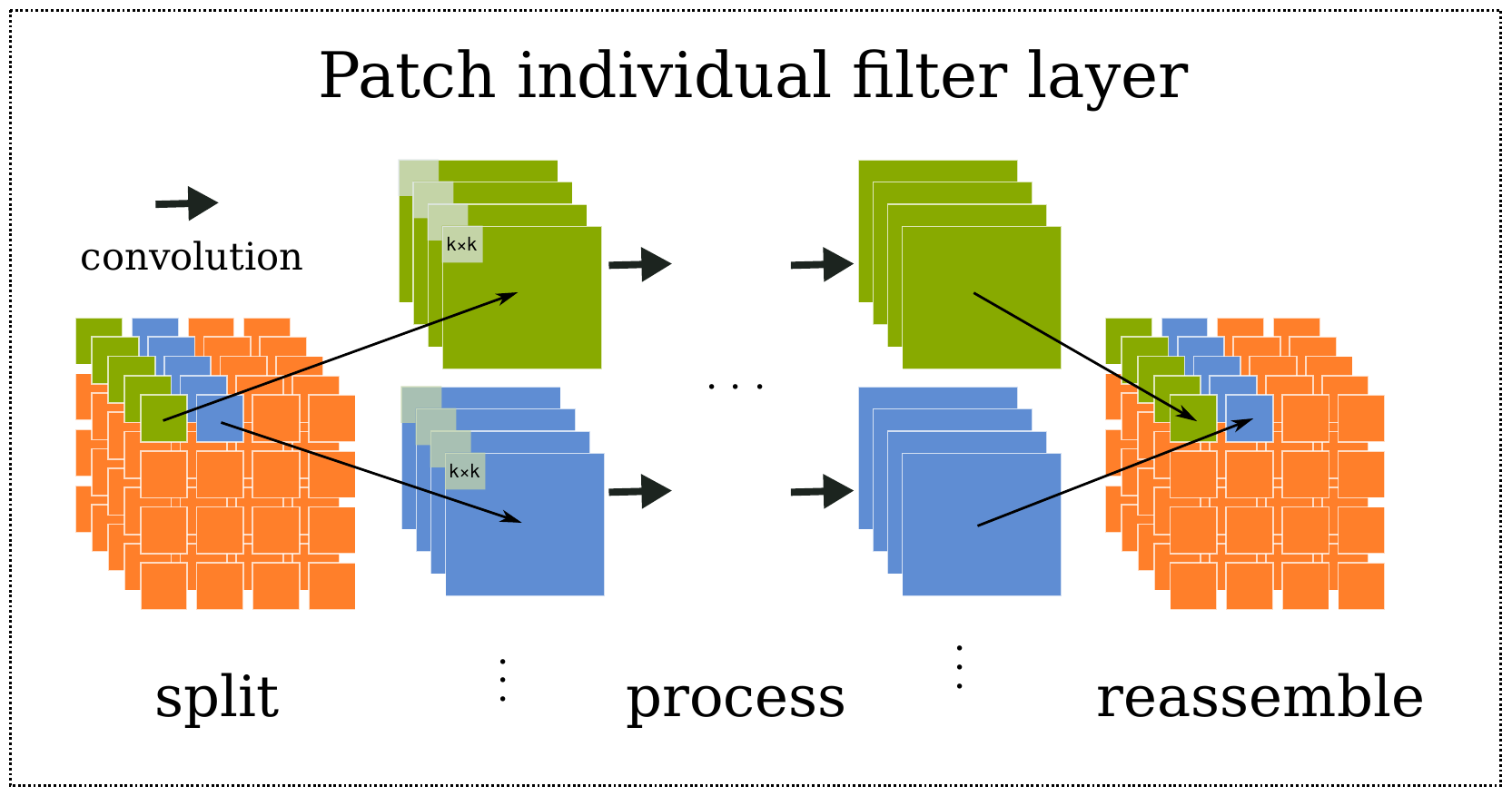}
    \caption{Depiction of a patch individual filter (PIF) layer in 2D. In this setting, inputs are 5 feature maps from a previous layer. Each feature map is being split into 16 patches and convolutions are applied patch-wise. Finally, the feature maps are reassembled in the same order.}
    \label{fig:layer}
\end{figure}

In comparison to the convolution operation in Equation \ref{eq:conv} in which a kernel \(K\) is convolved with an input \(I\), the PIF operation in Equation \ref{eq:pif} applies a patch specific kernel \(K_p\) to the current patch \(p\).
\begin{equation}\label{eq:conv}
    z = \sum_{m} \sum_{n} I(m,n)K(i-m,j-n),  
\end{equation}
\begin{equation}\label{eq:pif}
   z_p = \sum_{\hat{m}} \sum_{\hat{n}} I(\hat{m},\hat{n})K_p(i-\hat{m},j-\hat{n})\, \forall p \in P, 
\end{equation}
where \(\hat{m}, \hat{n} \in p\) and \(P\) is the set of all patches \(p\).
 When \(s > k\), weights are shared within each patch \(p_{ij}\)  but not across patches. Lastly, all patches are reassembled (iii) in the same order as they were split. Figure \ref{fig:layer} shows an overview of the layer design. PIF layers can be easily integrated into many CNN architectures and can be modified to contain other layer types besides convolutions. An implementation using PyTorch can be found here: \url{https://github.com/derEitel/patch_individual_filter_layer}.

When splitting a feature map into patches, one creates artificial borders which could reduce training performance. Each patch has several new and unnatural borders. These borders potentially cut through objects that the network might learn as a whole. For example the splitting could cause a feature map region representing the hippocampus to be split into two patches. The first downside is that this leads to potential border effects in areas that would normally not be affected. Second, a symptom such as hippocampal atrophy might only be visible in one of the patches, causing the two patches to disagree. Simply speaking, one patch might forward activations which support the disease class, while the other patch might inhibit activation, supporting the control class. To mitigate these issues we perform a parallel strain of network in which the patches are split with an overlap to the original split. Each original patch location \((x_o, y_o)\) is shifted by half its patch size to the overlapping location \((x_{ov}, y_{ov})\):

\begin{equation}\label{eq:overlap}
    (x_{ov}, y_{ov}) = (x_{o} + \frac{s_{x}}{2}, y_{o} + \frac{s_{y}}{2}).
\end{equation}
This way, patches are added in a minimalistic fashion, centralizing the overlap between existing patches while neglecting additional patches at the image borders that would require padding and are likely less informative.

\subsection{PIF layers in comparison}
\label{sec:comparison}

PIF layers are a novel concept but similar approaches exist and shall be presented here. PIF layers can be understood as a generalization of local convolutions as implemented in Lasagne\footnote{\url{https://lasagne.readthedocs.io/en/latest/modules/layers/local.html}} and Keras\footnote{\url{https://keras.io/layers/local/}}. Local convolutions are similar to regular convolutions but do not share weights across positions and are a special case of PIF layers where \(s + p = k\) with \(s\) being the patch size, \(p\) the padding size and \(k\) the kernel size. Thus, the convolution kernel does not slide over the selected patch (because they are congruent). 
In patch-based training \citep{Kamnitsas2016, Ghafoorian2017, Yoo2018}, multiple patches are sampled from the data set and fed into the same classifier regardless of the position of each patch. Since the filters of the classifier are applied on all patches, the weights are shared between patches. Conversely, within PIF layers, weights are only shared within a spatially restricted patch. Also, the main motivation for patch-based training, unlike for PIF layers, is to reduce the computational burden of high dimensional inputs.
PIF layers are furthermore different from PatchGANs \citep{li2016precomputed, isola2017image} which use Markovian patches as input for a discriminator network in order to focus penalization on high-frequency structures. 
Another approach introduced in \citet{kamyar2018Twostage} uses a greedy two-stage training strategy: first, a patch-wise model is trained, second the input image is split into 12 patches and latent features of the first model are extracted, and lastly those feature maps are concatenated to train a final classification network. Since the extracted feature maps are concatenated in order to create a spatially smaller 3D input for the classification network, weights are in turn shared between the feature map patches. 

\subsection{LRP Visualization}
\label{sec:lrp}
Layer-wise relevance propagation (LRP) is a method to analyze the behaviour of deep neural networks and other machine learning methods \citep{bach2015pixel}. It has been used in several studies in MR imaging \citep{yan2017LrpSz, Grigorescu2019InterpretableCN, islam2019understanding} and it was shown that identified relevant regions can overlap with clinically established relevant brain regions such as the medial temporal lobe in Alzheimer's disease \citep{Boehle2019, oh2019classification} and the corpus callosum in multiple sclerosis \citep{Eitel2019MS}. LRP uses backpropagation to transfer the output score of the network into the input space and therefore creates heatmaps which show the relevance of each pixel. We propagate the relevance following the \(\alpha\)/\(\beta\)-rule

\begin{equation}\label{eq:LRP}
    R_{i}^{l} = \sum_j \left( \alpha \frac{z_{ij}^{+}}{\sum_k z_{kj}^{+}} - \beta \frac{z_{ij}^{-}}{\sum_k z_{kj}^{-}} \right) R_{j}^{l+1}.
\end{equation}
Here, the relevance from layer \(R^{l+1}\) is backpropagated to its preceding layer \(R^{l}\). Activations are divided into positive and negative contributions (\(z^{+/-}_{ij})\) between nodes \(i\) and \(j\). Additionally, activations are normalized with the sum of the positive/negative activations from that layer. The hyperparameters \(\alpha\) and \(\beta\) need to be tuned and are confined to \(\alpha = 1 + \beta\). In this study we set \(\beta = 4\) as that was shown to be less prone to converging to a rank-1 matrix \citep{sixt2019explanations}. Typically, one invokes the LRP backpropagation with the activation of the final layer, here we furthermore start with the activation of hidden neurons in order to obtain the relevance of a specific filter within the network.

\section{Experiments}
\subsection{Data sets}
To study the effect of PIF layers we have compared the performance on three different structural MRI data sets. As we hypothesize that PIF layers should need fewer training examples to learn relevant features we have run an additional comparison on a randomly-sampled 20\% subset of the selected UK Biobank and ADNI sample. As the number of samples for the MS data set is already small, we did not use a subset here. 

\textbf{UK Biobank}
For this study, 1854 T1-weighted images (MPRAGE, 3 Tesla) from the UK Biobank\footnote{\url{https://www.ukbiobank.ac.uk/}} were randomly chosen to train a sex detection classifier. The MR images were obtained pre-processed from the UK Biobank repository combining data from several sites and scanners. The pre-processing pipeline included defacing, reduction of the field of view to remove empty space around the brain and gradient distortion correction. Furthermore, images were non-linearly transformed to MNI152 space using the FMRIB Software Library (FSL)\footnote{\url{https://fsl.fmrib.ox.ac.uk/fsl/fslwiki}}. The final image size is 182x218x182 voxels. The target of the trained classifiers is to distinguish between female (n = 1005) and male (n = 849) brains. Data was split into separate test (20\%), validation (16\%) and training (64\%) sets.

\textbf{ADNI}
969 T1-weighted images (MPRAGE, 1.5 Tesla) from the Alzheimer's Disease Neuroimaging Initiative (ADNI\footnote{\url{http://adni.loni.usc.edu/}}) database were used to discriminate subjects with Alzheimer's disease (AD) from healthy controls (HCs). The MR images come from different sites and scanners and were downloaded partially pre-processed. Already applied pre-processing steps included corrections for gradient non-linearity, intensity inhomogeneity and phantom-based distortion. We furthermore registered all images to the ICBM152 standard template (asymmetric version 2009c at 1 mm) using non-linear registration from the Advanced Normalization Tools (ANTs\footnote{\url{http://stnava.github.io/ANTs/}}). The final image size has been reduced to 96x114x96. The images stem from 193 AD patients and 151 HCs with up to three time points. To avoid data leakage, splitting of the data set was done on the patient level and not on the image level leading to disjoint test (18\%), validation (10\%) and training (72\%) sets.

\textbf{VIMS}
147 fluid-attenuated inversion recovery (FLAIR, 3 Tesla) images from the VIMS study\footnote{\url{https://neurocure.de/en/clinical-center/clinical-studies/current-studies.html}} of the NeuroCure center at Charité - Universitätsmedizin Berlin were used in order to separate patients with relapse-remitting multiple sclerosis (MS) and healthy controls (please see also \citet{Eitel2019MS}). All images were acquired from the same 3T scanner (Tim Trio Siemens, Erlangen, Germany). After bias-field correction and robust field of view selection, the corresponding MPRAGE sequences were linearly registered to MNI space using FSL\footnote{\url{https://fsl.fmrib.ox.ac.uk/fsl/fslwiki}}. FLAIR images were then co-registered to the MPRAGE images using spline interpolation. The final image size has been reduced to 96x114x96. 76 images stem from patients with relapsing-remitting multiple sclerosis (MS) according to the 2010 McDonald criteria \citep{polman2011diagnostic}, and the remaining 71 images stem from healthy controls. As the data set is small we dedicated a larger portion to the training set leading to splits of 15\% for testing, 8.5\% validation and 76.5\% training.

\subsection{Model architecture}

Based on the theoretical motivation, we compare here baseline CNN models, which were optimized in terms of architecture and hyperparameters for the given task, to the same CNN architecture in which the final convolutional layer (before fully-connected layers) was replaced with a PIF layer. 
 
 Since the PIF layers introduce additional features, each architecture was adjusted slightly to balance the total feature count between baseline and PIF model. This was mainly done to ensure that the proposed architecture does not profit from simply having a higher capacity than the baseline. In the following, we describe the final architectures for the three different tasks (sex classification on UK Biobank data, discrimination between patients with AD and healthy controls based on ADNI data and discrimination between patients with MS and healthy controls based on the VIMS data set). All models use 3D operations, have isotropic kernel sizes and were trained with binary cross entropy loss and the Adam optimizer. Before training, all models were randomly initialized using the initialization scheme described in \citet{he2015delving}.

\textbf{UK Biobank}
The Baseline-A network consists of five convolutional layers, each followed by an exponential linear unit (ELU) activation. MaxPooling and dropout with a probability of 30\% is applied after the first, second, and fifth layer. Each convolutional layer has a kernel of size 3 and the number of filters is 8, 16, 32, 64, 64 starting from the first layer. The pooling kernel is size 3 with stride 3 in the first two layers and kernel size 4 and stride 2 for the last layer. This is followed by two linear layers with 100 and 1 output features and a sigmoid activation. In the suggested PIF architecture, the final convolution and pooling layers have been replaced by a PIF layer. The PIF layer operates on patches of shape 5x5x5 with a kernel size 3. It consists of a single convolutional layer with 6 filters. Furthermore, overlapping patches were created with the same settings. For both the baseline and the PIF architecture the initial learning rate was set to 0.0001 and weight decay to 0.00001 on the full set, while the weight decay was increased to 0.0001 on the subset (20\% of the full set). Batch size was set to 8 for the baseline and 12 for the PIF as the latter has a smaller feature count. 

\textbf{ADNI}
Baseline-B is a slight modification of Baseline-A in which only the number of filters were changed to: 64, 64, 64, 64, 36 in the convolutional layers and 80, 1 in the fully-connected layers. The PIF architecture uses Baseline-B but replaces the fifth convolutional layer with a PIF layer. The PIF layer applies a single convolution on 5x5x5 patches with filter size 3 and 3 filters. Overlapping patches were created with the same settings. Training for all ADNI experiments was done using an initial learning rate of 0.0001 and a weight decay of 0.0001. We found the best batch size to be 12 for all models except PIF on the full data set where it was set to 6.

\textbf{VIMS}
The network for MS classification, Baseline-C is based on Baseline-A as well, but replaces the last convolutional layer with a  MaxPooling layer with additional dropout. It follows only a single fully-connected layer activated by a sigmoid function. The number of filters is changed to 64 in every remaining layer and MaxPooling is always applied using kernel size 3 and stride 3. The PIF architecture changes the filters of Baseline-C to 16, 32, 64, 64 and adds a PIF layer after the fourth convolutional layer, replacing the last MaxPooling layer and dropout operation. The PIF layer applies a single convolutional layer on each 5x5x5 patch with 4 filters. Overlapping patches were created with the same settings. Furthermore, a second fully-connected layer is added with 100 output features. Both models were trained using an initial learning rate of 0.0001 and weight decay of 0.0001. Batch size was set to 4.

\subsection{Data augmentation \& transformation}
For data augmentation, we used translation and flipping along the sagittal axis, which are typical augmentations in neuroimaging \citep{esmaeilzadeh2018end, eitel2019testing, basaia2019automated}. We applied data augmentation only to the UK Biobank and the ADNI data, but not to the VIMS data set, since here the validation performance was reduced when using data augmentation. For the PIF architecture, we performed only translation and not flipping, since the PIF layer requires patches to have the same content during each training iteration. 
All images were intensity normalized by dividing each by its maximum value before training. 


\subsection{Validation}
The main performance measure in our experiments was balanced accuracy, which is invariant with respect to unbalanced class distributions.  
Furthermore, to evaluate the hypothesis that PIF architectures require fewer training iterations, we additionally measured the number of iterations until early stopping occurs. Early stopping is a kind of regularization, which forces the model to end training after performance has not improved for a fixed amount of iterations. If a model converges quickly, it should obtain a high and stable performance early and therefore cause the early stopping algorithm to end training after fewer iterations than a slower converging model. Early stopping can also be caused by a model not being able to leave the initial optimization basin, or being stuck in a poor local minimum, hence using the number of early stopping iterations as a measure of convergence is only feasible when the model achieves a good optimization performance. Since performances of CNNs in the sample size and feature count settings of neuroimaging data tend to vary considerably \citep{eitel2019testing} we have repeated all experiments ten times with identical settings and report average results.\footnote{Sampling initial parameters is done separately for each repeat, all training parameters, sampling distributions and data splits remain identical.}


\section{Results and Discussion}
\label{sec:res}

Table \ref{tab:results} shows the results from all 5 comparisons averaged over 10 repeats. On the sex classification task (UK Biobank) the PIF model performs almost identical to the baseline on the full data set by gaining balanced accuracies of almost 94\% which is similar to the state-of-the art on structural MRI data \citep{Hu2019Sex, Anderson2019Sex, Schulz2019dlbrains}. Please note that \citet{Hu2019Sex} report their result (98.06\%) using 10-fold cross-validation instead of an exclusive test set which might inflate the results. \citet{Anderson2019Sex} report a cross-validation result similar to \citet{Hu2019Sex} (98.6\%) but it reduces to the level of our models (93.8\%) on a separate test set. Besides comparable performance, the number of iterations until early stopping reduce from 66.2 by almost a third when using the PIF architecture. On the smaller subset of only 371 samples our suggested architecture increases the balanced accuracy from 80.19\% by one-tenth to 88.41\%. Furthermore, the iteration in which early stopping occurs drops by one-third.
On the full ADNI data set the balanced accuracy increases slightly when using the PIF architecture from 83.60\% to 84.83\% when using all images and remains roughly identical to the baseline on the small data set around 79\%. In both cases the PIF architecture finishes training after just two-thirds of the training iterations. \citet{wen2020convolutional} is a recent and thorough overview of classification of AD on the ADNI data set along with reports of the issue of data leakage visible in many studies. The 13 reported studies in which no data leakage was found obtained  accuracies ranging from 76\% to 91\%. However, please note that 12 of those 13 studies report only non-balanced accuracy. 
In MS detection, the PIF architecture outperforms the baseline as well, increasing balanced accuracy from 76.35\% to 79.46\%. In \citet{Eitel2019MS}, a balanced accuracy of 71.23\% was reported using a CNN, which was improved to 87.04\% by using pre-training. \citet{Wang2018MS} report accuracies of almost 99\% using a comparably deep 14-layer CNN to detect MS. However, their study is based on two independent data sets in which the patients come from a different data generating distribution (different site, scanner, protocol) than the controls. The authors have solely used histogram matching to normalize image contrasts but not addressed any image differences beyond that. 
Besides, please note that most studies in MS using some variations of CNNs focus on the task of lesion segmentation \citep{VALVERDE2017159, KAMNITSAS2017lesions, nair2020exploring} rather than disease classification. 
Iterations until early stopping on the VIMS cohort drop on average from 80.7 to 60.6. 

 \begin{table*}[!t]
   \caption{Results of the three binary classification tasks. Balanced accuracies and iterations until early stopping are reported as averages over 10 repeated runs. Standard deviation (std) is reported in parentheses.}
   \label{tab:results}
   \centering
   \begin{tabular}{cccccc}
     \multicolumn{2}{c}{}
     & \multicolumn{2}{c}{\textbf{Large data set}}
     & \multicolumn{2}{c}{\textbf{Small data set}}
     \\
     Data & Model & Bal. acc. (std) & Early stopping iter. & Bal. acc. (std) & Early stopping iter. \\
     \hline
     UK Biobank & Baseline-A & 93.64\% (0.48) & 66.2 & 80.19\% (11.97) & 105.3 \\
     UK Biobank  & PIF-A & 93.75\% (1.09) & 48.9 & 88.41\% (3.52) & 72.7 \\
     \hline
     ADNI & Baseline-B & 83.60\% (4.50) & 34.2 & 79.71\% (3.23) & 106.8 \\
     ADNI & PIF-B & 84.83\% (1.65) & 21.3 & 79.47\% (2.23) & 73.1 \\
     \hline
     MS & Baseline-C & - & - & 76.35\% (8.09) & 80.7  \\
     MS & PIF-C & - & - & 79.46\% (9.50) & 60.6 \\
     \hline
   \end{tabular}
 \end{table*}

\subsection{Heatmap analysis}

We created several LRP heatmaps in order to support the motivation for PIF layers and determine differences between baseline and PIF model performances. Since the UK Biobank data set has the largest sample size and both baseline and the proposed architecture achieved a high, very similar performance, we only show visualizations for the UK Biobank data. For evaluation, a higher performance as well as a low difference in model performance is important. Otherwise, it remains unclear whether differences come from a gap in performance or the architecture itself. 

First, we used the baseline model to investigate whether higher layer features in a CNN trained on MRI data will have a more localized focus than features from lower layers. We generated LRP heatmaps using the outputs of both intermediary and the final output layer. Figure \ref{fig:lrp_bsl} shows the heatmaps obtained by backpropagating the activations of 4 randomly selected filters in convolutional layers 4 and 5 as well as the heatmap of the model output. The comparison between the heatmaps of convolutional layers 4 and 5 shows that the lower layer has more connected and dense heatmaps across all shown filters whereas the higher layers have more sparse heatmaps with several regions showing no activity, such as the frontal left hemisphere in the filter 0 heatmap, the cortex of the right hemisphere of the filter 15 heatmap or posterior and anterior regions in the center of the right hemisphere in the filter 60 heatmap. The LRP heatmap of the model score on the other hand does not portray these empty regions. As the final output is a combination of all convolutional filters the heatmap becomes more holistic and dense again. Nevertheless, trends emerge in many layer 5 filters which are reflected in the final output, such as a low focus on the inferior frontal gyrus. This shows that the filters of the last convolutional layer tend to be more locally specific in the baseline architecture. As the PIF architecture enforces this higher layer locality through its feature map patches, it could be one of the causes of its faster convergence over the baseline.  
 \begin{figure*}
    \centering
    \input{figures/LRP_heatmaps_baseline_UKB.pgf}
    \caption{Heatmaps of the UK Biobank baseline (Baseline-A) generated from the last two convolutional layers and the final output. Four out of the 64 filters for the convolutional layers were randomly selected. Note that there is no special relation between the filters at the same location (i.e. filter 0 at conv 4 and conv 5) as each filter is applied to all previous feature maps.}
    \label{fig:lrp_bsl}
\end{figure*}
 
Next, we compared heatmaps from the baseline model to the proposed PIF architecture. Figure \ref{fig:lrp_patch} shows the heatmaps of the UK Biobank PIF architecture, obtained from the PIF layer. Here, the locality due to the patches is highly apparent. Patches 1 to 36 are the original patches, whereas patches 37 to 48 are the overlapping patches. One can see that patches 40 and 41 overlay regions of patch 10 and neighboring filters are slightly translated such as patches 40 and 41. Beyond that, the layer 4 heatmaps of the PIF model, as depicted in Figure \ref{fig:lrp_pif_conv4+score}, show that the sparseness occurring in baseline layer 5 does not simply move down to layer 4 in the PIF model. If this had been the case, it could have been an indicator that the PIF layer does not add any disentanglement value and that the remaining model is able to learn similar features with a smaller capacity. Here, we can rather see that the layer 4 heatmaps between baseline and PIF model seem to be very similar in terms of general structure, indicating that the PIF layer does not obstruct the general learning performance of the CNN. Finally, the LRP heatmap of the final output in Figure \ref{fig:lrp_pif_conv4+score} shows also strong resemblance to the baseline heatmap in Figure \ref{fig:lrp_bsl}.
 
 \begin{figure*}
    \centering
    \input{figures/LRP_heatmaps_experiment_PIF-layer_UKB.pgf}
    \caption{LRP heatmaps of the UK Biobank PIF architecture (PIF-A) using the PIF layer output to generate patch and filter specific heatmaps. Each patch learns individual filters and therefore patches at the same location (i.e. filter 0 across patches) do not share a specific relation.}
    \label{fig:lrp_patch}
\end{figure*}
 
 \begin{figure*}
    \centering
    \input{figures/LRP_heatmaps_experiment_conv4+score_UKB.pgf}
    \caption{LRP heatmaps of PIF-A based on layer 4 feature maps (the final layer before the PIF layer) and the model output (score).}
    \label{fig:lrp_pif_conv4+score}
\end{figure*}

\subsection{Limitations}
The major limitation of PIF layers is that it requires all examples to have the same dimensions. By design PIF layers require spatial standardization and inputs need to have the same number of features per dimension (e.g. same number of voxels). As the pre-processing was optimized for each data set individually, this refrained us from training a single classifier on all presented tasks. Nevertheless, when MRI data as well as pre-processing pipelines become more and more standardized, a holistic architecture might become conceivable. 
A second limitation are potential border effects.  
While we reduced the risk of splitting important objects by using overlapping patches, we thereby introduced additional artificial borders which might be unfavorable and their effect should be addressed in future studies. 
And finally, due to many ways of how to pre-process and split the data as well as large hyperparameter spaces, it is generally difficult to compare different deep learning algorithms and thus we can not rule out that better configurations for each of the tasks exist. However, the employed PIF architectures show improvements on all 6 experiments over successful baselines. To keep the comparison as simple as possible, we only used randomly initialized CNN baselines. Future studies might investigate the effect of transfer learning, combining several modalities and other validation strategies.

\section{Conclusion}
\label{sec:conc}
In this study, we have introduced PIF layers for CNNs. Based on the understanding that higher level layers learn more abstract and localized features, we have reinforced that learning direction by splitting higher level feature maps into patches and learning features without weight-sharing between patches. In scenarios, where data is naturally homogeneous or spatially normalized, PIF layers can be introduced in order to reduce the number of samples and iterations for model convergence. This method can be seen as introducing a spatial prior into the neural network model. 
Based on further knowledge about the data, one could tune this prior by adjusting the patch size to the size of a certain biomarker or relevant sub-regions in an image, or could weigh patches based on a pre-defined hypothesis.
Here, we have shown that PIF layers can be beneficial for classification based on human brain MRI. 
Potential future applications are other standardized medical sets, e.g., coming from other modalities (PET, CT, other MRI sequences etc.) or other parts of the body also requiring normalization.



\section{CRediT author statement}
\textbf{Faban Eitel:} Conceptualization, Methodology, Writing - Original Draft, Software
\textbf{Jan Philipp Albrecht:} Formal analysis, Methodology, Software, Investigation 
\textbf{Martin Weygandt:} Writing - Review \& Editing 
\textbf{Friedemann Paul:} Supervision, Resources
\textbf{Kerstin Ritter:} Conceptualization, Supervision, Writing - Review \& Editing

\section{Acknowledgements}
We acknowledge support from the German Research Foundation (DFG, 389563835), the Deutsche Multiple Sklerose Gesellschaft (DMSG) Bundesverband e.V., the Manfred and Ursula-Müller Stiftung and Charité – Universitätsmedizin Berlin (Open Access Publication Fund).

\bibliographystyle{model2-names.bst}
\bibliography{refs}

\end{document}

%% file: figures/LRP_heatmaps_baseline_UKB.pgf
\begingroup%
\makeatletter%
\begin{pgfpicture}%
\pgfpathrectangle{\pgfpointorigin}{\pgfqpoint{6.065323in}{3.322969in}}%
\pgfusepath{use as bounding box, clip}%
\begin{pgfscope}%
\pgfsetbuttcap%
\pgfsetmiterjoin%
\definecolor{currentfill}{rgb}{1.000000,1.000000,1.000000}%
\pgfsetfillcolor{currentfill}%
\pgfsetlinewidth{0.000000pt}%
\definecolor{currentstroke}{rgb}{1.000000,1.000000,1.000000}%
\pgfsetstrokecolor{currentstroke}%
\pgfsetdash{}{0pt}%
\pgfpathmoveto{\pgfqpoint{0.000000in}{-0.000000in}}%
\pgfpathlineto{\pgfqpoint{6.065323in}{-0.000000in}}%
\pgfpathlineto{\pgfqpoint{6.065323in}{3.322969in}}%
\pgfpathlineto{\pgfqpoint{0.000000in}{3.322969in}}%
\pgfpathclose%
\pgfusepath{fill}%
\end{pgfscope}%
\begin{pgfscope}%
\definecolor{textcolor}{rgb}{0.000000,0.000000,0.000000}%
\pgfsetstrokecolor{textcolor}%
\pgfsetfillcolor{textcolor}%
\pgftext[x=1.223581in,y=3.126518in,,base]{\color{textcolor}\rmfamily\fontsize{9.600000}{11.520000}\bfseries\selectfont Layer Conv4}%
\end{pgfscope}%
\begin{pgfscope}%
\pgfsetbuttcap%
\pgfsetmiterjoin%
\definecolor{currentfill}{rgb}{1.000000,1.000000,1.000000}%
\pgfsetfillcolor{currentfill}%
\pgfsetlinewidth{0.000000pt}%
\definecolor{currentstroke}{rgb}{0.000000,0.000000,0.000000}%
\pgfsetstrokecolor{currentstroke}%
\pgfsetstrokeopacity{0.000000}%
\pgfsetdash{}{0pt}%
\pgfpathmoveto{\pgfqpoint{0.348073in}{1.896668in}}%
\pgfpathlineto{\pgfqpoint{1.181890in}{1.896668in}}%
\pgfpathlineto{\pgfqpoint{1.181890in}{2.895416in}}%
\pgfpathlineto{\pgfqpoint{0.348073in}{2.895416in}}%
\pgfpathclose%
\pgfusepath{fill}%
\end{pgfscope}%
\begin{pgfscope}%
\pgfpathrectangle{\pgfqpoint{0.348073in}{1.896668in}}{\pgfqpoint{0.833817in}{0.998748in}}%
\pgfusepath{clip}%
\pgfsys@transformshift{0.348073in}{1.896668in}%
\pgftext[left,bottom]{\pgfimage[interpolate=true,width=0.836000in,height=1.000000in]{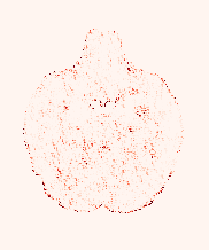}}%
\end{pgfscope}%
\begin{pgfscope}%
\pgfsetrectcap%
\pgfsetmiterjoin%
\pgfsetlinewidth{0.803000pt}%
\definecolor{currentstroke}{rgb}{0.000000,0.000000,0.000000}%
\pgfsetstrokecolor{currentstroke}%
\pgfsetdash{}{0pt}%
\pgfpathmoveto{\pgfqpoint{0.348073in}{1.896668in}}%
\pgfpathlineto{\pgfqpoint{0.348073in}{2.895416in}}%
\pgfusepath{stroke}%
\end{pgfscope}%
\begin{pgfscope}%
\pgfsetrectcap%
\pgfsetmiterjoin%
\pgfsetlinewidth{0.803000pt}%
\definecolor{currentstroke}{rgb}{0.000000,0.000000,0.000000}%
\pgfsetstrokecolor{currentstroke}%
\pgfsetdash{}{0pt}%
\pgfpathmoveto{\pgfqpoint{1.181890in}{1.896668in}}%
\pgfpathlineto{\pgfqpoint{1.181890in}{2.895416in}}%
\pgfusepath{stroke}%
\end{pgfscope}%
\begin{pgfscope}%
\pgfsetrectcap%
\pgfsetmiterjoin%
\pgfsetlinewidth{0.803000pt}%
\definecolor{currentstroke}{rgb}{0.000000,0.000000,0.000000}%
\pgfsetstrokecolor{currentstroke}%
\pgfsetdash{}{0pt}%
\pgfpathmoveto{\pgfqpoint{0.348073in}{1.896668in}}%
\pgfpathlineto{\pgfqpoint{1.181890in}{1.896668in}}%
\pgfusepath{stroke}%
\end{pgfscope}%
\begin{pgfscope}%
\pgfsetrectcap%
\pgfsetmiterjoin%
\pgfsetlinewidth{0.803000pt}%
\definecolor{currentstroke}{rgb}{0.000000,0.000000,0.000000}%
\pgfsetstrokecolor{currentstroke}%
\pgfsetdash{}{0pt}%
\pgfpathmoveto{\pgfqpoint{0.348073in}{2.895416in}}%
\pgfpathlineto{\pgfqpoint{1.181890in}{2.895416in}}%
\pgfusepath{stroke}%
\end{pgfscope}%
\begin{pgfscope}%
\definecolor{textcolor}{rgb}{0.000000,0.000000,0.000000}%
\pgfsetstrokecolor{textcolor}%
\pgfsetfillcolor{textcolor}%
\pgftext[x=0.764982in,y=2.978750in,,base]{\color{textcolor}\rmfamily\fontsize{9.600000}{11.520000}\selectfont Filter 0}%
\end{pgfscope}%
\begin{pgfscope}%
\pgfsetbuttcap%
\pgfsetmiterjoin%
\definecolor{currentfill}{rgb}{1.000000,1.000000,1.000000}%
\pgfsetfillcolor{currentfill}%
\pgfsetlinewidth{0.000000pt}%
\definecolor{currentstroke}{rgb}{0.000000,0.000000,0.000000}%
\pgfsetstrokecolor{currentstroke}%
\pgfsetstrokeopacity{0.000000}%
\pgfsetdash{}{0pt}%
\pgfpathmoveto{\pgfqpoint{1.265272in}{1.896668in}}%
\pgfpathlineto{\pgfqpoint{2.099089in}{1.896668in}}%
\pgfpathlineto{\pgfqpoint{2.099089in}{2.895416in}}%
\pgfpathlineto{\pgfqpoint{1.265272in}{2.895416in}}%
\pgfpathclose%
\pgfusepath{fill}%
\end{pgfscope}%
\begin{pgfscope}%
\pgfpathrectangle{\pgfqpoint{1.265272in}{1.896668in}}{\pgfqpoint{0.833817in}{0.998748in}}%
\pgfusepath{clip}%
\pgfsys@transformshift{1.265272in}{1.896668in}%
\pgftext[left,bottom]{\pgfimage[interpolate=true,width=0.836000in,height=1.000000in]{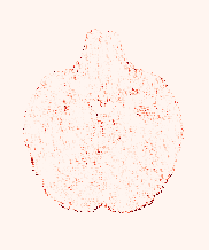}}%
\end{pgfscope}%
\begin{pgfscope}%
\pgfsetrectcap%
\pgfsetmiterjoin%
\pgfsetlinewidth{0.803000pt}%
\definecolor{currentstroke}{rgb}{0.000000,0.000000,0.000000}%
\pgfsetstrokecolor{currentstroke}%
\pgfsetdash{}{0pt}%
\pgfpathmoveto{\pgfqpoint{1.265272in}{1.896668in}}%
\pgfpathlineto{\pgfqpoint{1.265272in}{2.895416in}}%
\pgfusepath{stroke}%
\end{pgfscope}%
\begin{pgfscope}%
\pgfsetrectcap%
\pgfsetmiterjoin%
\pgfsetlinewidth{0.803000pt}%
\definecolor{currentstroke}{rgb}{0.000000,0.000000,0.000000}%
\pgfsetstrokecolor{currentstroke}%
\pgfsetdash{}{0pt}%
\pgfpathmoveto{\pgfqpoint{2.099089in}{1.896668in}}%
\pgfpathlineto{\pgfqpoint{2.099089in}{2.895416in}}%
\pgfusepath{stroke}%
\end{pgfscope}%
\begin{pgfscope}%
\pgfsetrectcap%
\pgfsetmiterjoin%
\pgfsetlinewidth{0.803000pt}%
\definecolor{currentstroke}{rgb}{0.000000,0.000000,0.000000}%
\pgfsetstrokecolor{currentstroke}%
\pgfsetdash{}{0pt}%
\pgfpathmoveto{\pgfqpoint{1.265272in}{1.896668in}}%
\pgfpathlineto{\pgfqpoint{2.099089in}{1.896668in}}%
\pgfusepath{stroke}%
\end{pgfscope}%
\begin{pgfscope}%
\pgfsetrectcap%
\pgfsetmiterjoin%
\pgfsetlinewidth{0.803000pt}%
\definecolor{currentstroke}{rgb}{0.000000,0.000000,0.000000}%
\pgfsetstrokecolor{currentstroke}%
\pgfsetdash{}{0pt}%
\pgfpathmoveto{\pgfqpoint{1.265272in}{2.895416in}}%
\pgfpathlineto{\pgfqpoint{2.099089in}{2.895416in}}%
\pgfusepath{stroke}%
\end{pgfscope}%
\begin{pgfscope}%
\definecolor{textcolor}{rgb}{0.000000,0.000000,0.000000}%
\pgfsetstrokecolor{textcolor}%
\pgfsetfillcolor{textcolor}%
\pgftext[x=1.682180in,y=2.978750in,,base]{\color{textcolor}\rmfamily\fontsize{9.600000}{11.520000}\selectfont Filter 4}%
\end{pgfscope}%
\begin{pgfscope}%
\pgfsetbuttcap%
\pgfsetmiterjoin%
\definecolor{currentfill}{rgb}{1.000000,1.000000,1.000000}%
\pgfsetfillcolor{currentfill}%
\pgfsetlinewidth{0.000000pt}%
\definecolor{currentstroke}{rgb}{0.000000,0.000000,0.000000}%
\pgfsetstrokecolor{currentstroke}%
\pgfsetstrokeopacity{0.000000}%
\pgfsetdash{}{0pt}%
\pgfpathmoveto{\pgfqpoint{0.348073in}{0.472954in}}%
\pgfpathlineto{\pgfqpoint{1.181890in}{0.472954in}}%
\pgfpathlineto{\pgfqpoint{1.181890in}{1.471702in}}%
\pgfpathlineto{\pgfqpoint{0.348073in}{1.471702in}}%
\pgfpathclose%
\pgfusepath{fill}%
\end{pgfscope}%
\begin{pgfscope}%
\pgfpathrectangle{\pgfqpoint{0.348073in}{0.472954in}}{\pgfqpoint{0.833817in}{0.998748in}}%
\pgfusepath{clip}%
\pgfsys@transformshift{0.348073in}{0.472954in}%
\pgftext[left,bottom]{\pgfimage[interpolate=true,width=0.836000in,height=1.000000in]{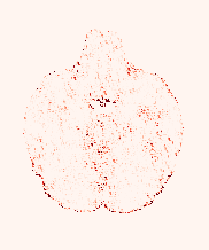}}%
\end{pgfscope}%
\begin{pgfscope}%
\pgfsetrectcap%
\pgfsetmiterjoin%
\pgfsetlinewidth{0.803000pt}%
\definecolor{currentstroke}{rgb}{0.000000,0.000000,0.000000}%
\pgfsetstrokecolor{currentstroke}%
\pgfsetdash{}{0pt}%
\pgfpathmoveto{\pgfqpoint{0.348073in}{0.472954in}}%
\pgfpathlineto{\pgfqpoint{0.348073in}{1.471702in}}%
\pgfusepath{stroke}%
\end{pgfscope}%
\begin{pgfscope}%
\pgfsetrectcap%
\pgfsetmiterjoin%
\pgfsetlinewidth{0.803000pt}%
\definecolor{currentstroke}{rgb}{0.000000,0.000000,0.000000}%
\pgfsetstrokecolor{currentstroke}%
\pgfsetdash{}{0pt}%
\pgfpathmoveto{\pgfqpoint{1.181890in}{0.472954in}}%
\pgfpathlineto{\pgfqpoint{1.181890in}{1.471702in}}%
\pgfusepath{stroke}%
\end{pgfscope}%
\begin{pgfscope}%
\pgfsetrectcap%
\pgfsetmiterjoin%
\pgfsetlinewidth{0.803000pt}%
\definecolor{currentstroke}{rgb}{0.000000,0.000000,0.000000}%
\pgfsetstrokecolor{currentstroke}%
\pgfsetdash{}{0pt}%
\pgfpathmoveto{\pgfqpoint{0.348073in}{0.472954in}}%
\pgfpathlineto{\pgfqpoint{1.181890in}{0.472954in}}%
\pgfusepath{stroke}%
\end{pgfscope}%
\begin{pgfscope}%
\pgfsetrectcap%
\pgfsetmiterjoin%
\pgfsetlinewidth{0.803000pt}%
\definecolor{currentstroke}{rgb}{0.000000,0.000000,0.000000}%
\pgfsetstrokecolor{currentstroke}%
\pgfsetdash{}{0pt}%
\pgfpathmoveto{\pgfqpoint{0.348073in}{1.471702in}}%
\pgfpathlineto{\pgfqpoint{1.181890in}{1.471702in}}%
\pgfusepath{stroke}%
\end{pgfscope}%
\begin{pgfscope}%
\definecolor{textcolor}{rgb}{0.000000,0.000000,0.000000}%
\pgfsetstrokecolor{textcolor}%
\pgfsetfillcolor{textcolor}%
\pgftext[x=0.764982in,y=1.555035in,,base]{\color{textcolor}\rmfamily\fontsize{9.600000}{11.520000}\selectfont Filter 15}%
\end{pgfscope}%
\begin{pgfscope}%
\pgfsetbuttcap%
\pgfsetmiterjoin%
\definecolor{currentfill}{rgb}{1.000000,1.000000,1.000000}%
\pgfsetfillcolor{currentfill}%
\pgfsetlinewidth{0.000000pt}%
\definecolor{currentstroke}{rgb}{0.000000,0.000000,0.000000}%
\pgfsetstrokecolor{currentstroke}%
\pgfsetstrokeopacity{0.000000}%
\pgfsetdash{}{0pt}%
\pgfpathmoveto{\pgfqpoint{1.265272in}{0.472954in}}%
\pgfpathlineto{\pgfqpoint{2.099089in}{0.472954in}}%
\pgfpathlineto{\pgfqpoint{2.099089in}{1.471702in}}%
\pgfpathlineto{\pgfqpoint{1.265272in}{1.471702in}}%
\pgfpathclose%
\pgfusepath{fill}%
\end{pgfscope}%
\begin{pgfscope}%
\pgfpathrectangle{\pgfqpoint{1.265272in}{0.472954in}}{\pgfqpoint{0.833817in}{0.998748in}}%
\pgfusepath{clip}%
\pgfsys@transformshift{1.265272in}{0.472954in}%
\pgftext[left,bottom]{\pgfimage[interpolate=true,width=0.836000in,height=1.000000in]{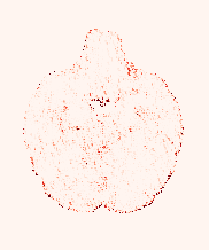}}%
\end{pgfscope}%
\begin{pgfscope}%
\pgfsetrectcap%
\pgfsetmiterjoin%
\pgfsetlinewidth{0.803000pt}%
\definecolor{currentstroke}{rgb}{0.000000,0.000000,0.000000}%
\pgfsetstrokecolor{currentstroke}%
\pgfsetdash{}{0pt}%
\pgfpathmoveto{\pgfqpoint{1.265272in}{0.472954in}}%
\pgfpathlineto{\pgfqpoint{1.265272in}{1.471702in}}%
\pgfusepath{stroke}%
\end{pgfscope}%
\begin{pgfscope}%
\pgfsetrectcap%
\pgfsetmiterjoin%
\pgfsetlinewidth{0.803000pt}%
\definecolor{currentstroke}{rgb}{0.000000,0.000000,0.000000}%
\pgfsetstrokecolor{currentstroke}%
\pgfsetdash{}{0pt}%
\pgfpathmoveto{\pgfqpoint{2.099089in}{0.472954in}}%
\pgfpathlineto{\pgfqpoint{2.099089in}{1.471702in}}%
\pgfusepath{stroke}%
\end{pgfscope}%
\begin{pgfscope}%
\pgfsetrectcap%
\pgfsetmiterjoin%
\pgfsetlinewidth{0.803000pt}%
\definecolor{currentstroke}{rgb}{0.000000,0.000000,0.000000}%
\pgfsetstrokecolor{currentstroke}%
\pgfsetdash{}{0pt}%
\pgfpathmoveto{\pgfqpoint{1.265272in}{0.472954in}}%
\pgfpathlineto{\pgfqpoint{2.099089in}{0.472954in}}%
\pgfusepath{stroke}%
\end{pgfscope}%
\begin{pgfscope}%
\pgfsetrectcap%
\pgfsetmiterjoin%
\pgfsetlinewidth{0.803000pt}%
\definecolor{currentstroke}{rgb}{0.000000,0.000000,0.000000}%
\pgfsetstrokecolor{currentstroke}%
\pgfsetdash{}{0pt}%
\pgfpathmoveto{\pgfqpoint{1.265272in}{1.471702in}}%
\pgfpathlineto{\pgfqpoint{2.099089in}{1.471702in}}%
\pgfusepath{stroke}%
\end{pgfscope}%
\begin{pgfscope}%
\definecolor{textcolor}{rgb}{0.000000,0.000000,0.000000}%
\pgfsetstrokecolor{textcolor}%
\pgfsetfillcolor{textcolor}%
\pgftext[x=1.682180in,y=1.555035in,,base]{\color{textcolor}\rmfamily\fontsize{9.600000}{11.520000}\selectfont Filter 60}%
\end{pgfscope}%
\begin{pgfscope}%
\definecolor{textcolor}{rgb}{0.000000,0.000000,0.000000}%
\pgfsetstrokecolor{textcolor}%
\pgfsetfillcolor{textcolor}%
\pgftext[x=3.149698in,y=3.126518in,,base]{\color{textcolor}\rmfamily\fontsize{9.600000}{11.520000}\bfseries\selectfont Layer Conv5}%
\end{pgfscope}%
\begin{pgfscope}%
\pgfsetbuttcap%
\pgfsetmiterjoin%
\definecolor{currentfill}{rgb}{1.000000,1.000000,1.000000}%
\pgfsetfillcolor{currentfill}%
\pgfsetlinewidth{0.000000pt}%
\definecolor{currentstroke}{rgb}{0.000000,0.000000,0.000000}%
\pgfsetstrokecolor{currentstroke}%
\pgfsetstrokeopacity{0.000000}%
\pgfsetdash{}{0pt}%
\pgfpathmoveto{\pgfqpoint{2.274190in}{1.896668in}}%
\pgfpathlineto{\pgfqpoint{3.108007in}{1.896668in}}%
\pgfpathlineto{\pgfqpoint{3.108007in}{2.895416in}}%
\pgfpathlineto{\pgfqpoint{2.274190in}{2.895416in}}%
\pgfpathclose%
\pgfusepath{fill}%
\end{pgfscope}%
\begin{pgfscope}%
\pgfpathrectangle{\pgfqpoint{2.274190in}{1.896668in}}{\pgfqpoint{0.833817in}{0.998748in}}%
\pgfusepath{clip}%
\pgfsys@transformshift{2.274190in}{1.896668in}%
\pgftext[left,bottom]{\pgfimage[interpolate=true,width=0.836000in,height=1.000000in]{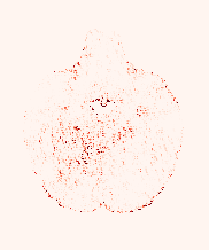}}%
\end{pgfscope}%
\begin{pgfscope}%
\pgfsetrectcap%
\pgfsetmiterjoin%
\pgfsetlinewidth{0.803000pt}%
\definecolor{currentstroke}{rgb}{0.000000,0.000000,0.000000}%
\pgfsetstrokecolor{currentstroke}%
\pgfsetdash{}{0pt}%
\pgfpathmoveto{\pgfqpoint{2.274190in}{1.896668in}}%
\pgfpathlineto{\pgfqpoint{2.274190in}{2.895416in}}%
\pgfusepath{stroke}%
\end{pgfscope}%
\begin{pgfscope}%
\pgfsetrectcap%
\pgfsetmiterjoin%
\pgfsetlinewidth{0.803000pt}%
\definecolor{currentstroke}{rgb}{0.000000,0.000000,0.000000}%
\pgfsetstrokecolor{currentstroke}%
\pgfsetdash{}{0pt}%
\pgfpathmoveto{\pgfqpoint{3.108007in}{1.896668in}}%
\pgfpathlineto{\pgfqpoint{3.108007in}{2.895416in}}%
\pgfusepath{stroke}%
\end{pgfscope}%
\begin{pgfscope}%
\pgfsetrectcap%
\pgfsetmiterjoin%
\pgfsetlinewidth{0.803000pt}%
\definecolor{currentstroke}{rgb}{0.000000,0.000000,0.000000}%
\pgfsetstrokecolor{currentstroke}%
\pgfsetdash{}{0pt}%
\pgfpathmoveto{\pgfqpoint{2.274190in}{1.896668in}}%
\pgfpathlineto{\pgfqpoint{3.108007in}{1.896668in}}%
\pgfusepath{stroke}%
\end{pgfscope}%
\begin{pgfscope}%
\pgfsetrectcap%
\pgfsetmiterjoin%
\pgfsetlinewidth{0.803000pt}%
\definecolor{currentstroke}{rgb}{0.000000,0.000000,0.000000}%
\pgfsetstrokecolor{currentstroke}%
\pgfsetdash{}{0pt}%
\pgfpathmoveto{\pgfqpoint{2.274190in}{2.895416in}}%
\pgfpathlineto{\pgfqpoint{3.108007in}{2.895416in}}%
\pgfusepath{stroke}%
\end{pgfscope}%
\begin{pgfscope}%
\definecolor{textcolor}{rgb}{0.000000,0.000000,0.000000}%
\pgfsetstrokecolor{textcolor}%
\pgfsetfillcolor{textcolor}%
\pgftext[x=2.691099in,y=2.978750in,,base]{\color{textcolor}\rmfamily\fontsize{9.600000}{11.520000}\selectfont Filter 0}%
\end{pgfscope}%
\begin{pgfscope}%
\pgfsetbuttcap%
\pgfsetmiterjoin%
\definecolor{currentfill}{rgb}{1.000000,1.000000,1.000000}%
\pgfsetfillcolor{currentfill}%
\pgfsetlinewidth{0.000000pt}%
\definecolor{currentstroke}{rgb}{0.000000,0.000000,0.000000}%
\pgfsetstrokecolor{currentstroke}%
\pgfsetstrokeopacity{0.000000}%
\pgfsetdash{}{0pt}%
\pgfpathmoveto{\pgfqpoint{3.191389in}{1.896668in}}%
\pgfpathlineto{\pgfqpoint{4.025206in}{1.896668in}}%
\pgfpathlineto{\pgfqpoint{4.025206in}{2.895416in}}%
\pgfpathlineto{\pgfqpoint{3.191389in}{2.895416in}}%
\pgfpathclose%
\pgfusepath{fill}%
\end{pgfscope}%
\begin{pgfscope}%
\pgfpathrectangle{\pgfqpoint{3.191389in}{1.896668in}}{\pgfqpoint{0.833817in}{0.998748in}}%
\pgfusepath{clip}%
\pgfsys@transformshift{3.191389in}{1.896668in}%
\pgftext[left,bottom]{\pgfimage[interpolate=true,width=0.836000in,height=1.000000in]{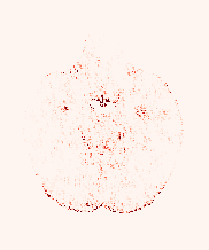}}%
\end{pgfscope}%
\begin{pgfscope}%
\pgfsetrectcap%
\pgfsetmiterjoin%
\pgfsetlinewidth{0.803000pt}%
\definecolor{currentstroke}{rgb}{0.000000,0.000000,0.000000}%
\pgfsetstrokecolor{currentstroke}%
\pgfsetdash{}{0pt}%
\pgfpathmoveto{\pgfqpoint{3.191389in}{1.896668in}}%
\pgfpathlineto{\pgfqpoint{3.191389in}{2.895416in}}%
\pgfusepath{stroke}%
\end{pgfscope}%
\begin{pgfscope}%
\pgfsetrectcap%
\pgfsetmiterjoin%
\pgfsetlinewidth{0.803000pt}%
\definecolor{currentstroke}{rgb}{0.000000,0.000000,0.000000}%
\pgfsetstrokecolor{currentstroke}%
\pgfsetdash{}{0pt}%
\pgfpathmoveto{\pgfqpoint{4.025206in}{1.896668in}}%
\pgfpathlineto{\pgfqpoint{4.025206in}{2.895416in}}%
\pgfusepath{stroke}%
\end{pgfscope}%
\begin{pgfscope}%
\pgfsetrectcap%
\pgfsetmiterjoin%
\pgfsetlinewidth{0.803000pt}%
\definecolor{currentstroke}{rgb}{0.000000,0.000000,0.000000}%
\pgfsetstrokecolor{currentstroke}%
\pgfsetdash{}{0pt}%
\pgfpathmoveto{\pgfqpoint{3.191389in}{1.896668in}}%
\pgfpathlineto{\pgfqpoint{4.025206in}{1.896668in}}%
\pgfusepath{stroke}%
\end{pgfscope}%
\begin{pgfscope}%
\pgfsetrectcap%
\pgfsetmiterjoin%
\pgfsetlinewidth{0.803000pt}%
\definecolor{currentstroke}{rgb}{0.000000,0.000000,0.000000}%
\pgfsetstrokecolor{currentstroke}%
\pgfsetdash{}{0pt}%
\pgfpathmoveto{\pgfqpoint{3.191389in}{2.895416in}}%
\pgfpathlineto{\pgfqpoint{4.025206in}{2.895416in}}%
\pgfusepath{stroke}%
\end{pgfscope}%
\begin{pgfscope}%
\definecolor{textcolor}{rgb}{0.000000,0.000000,0.000000}%
\pgfsetstrokecolor{textcolor}%
\pgfsetfillcolor{textcolor}%
\pgftext[x=3.608298in,y=2.978750in,,base]{\color{textcolor}\rmfamily\fontsize{9.600000}{11.520000}\selectfont Filter 4}%
\end{pgfscope}%
\begin{pgfscope}%
\pgfsetbuttcap%
\pgfsetmiterjoin%
\definecolor{currentfill}{rgb}{1.000000,1.000000,1.000000}%
\pgfsetfillcolor{currentfill}%
\pgfsetlinewidth{0.000000pt}%
\definecolor{currentstroke}{rgb}{0.000000,0.000000,0.000000}%
\pgfsetstrokecolor{currentstroke}%
\pgfsetstrokeopacity{0.000000}%
\pgfsetdash{}{0pt}%
\pgfpathmoveto{\pgfqpoint{2.274190in}{0.472954in}}%
\pgfpathlineto{\pgfqpoint{3.108007in}{0.472954in}}%
\pgfpathlineto{\pgfqpoint{3.108007in}{1.471702in}}%
\pgfpathlineto{\pgfqpoint{2.274190in}{1.471702in}}%
\pgfpathclose%
\pgfusepath{fill}%
\end{pgfscope}%
\begin{pgfscope}%
\pgfpathrectangle{\pgfqpoint{2.274190in}{0.472954in}}{\pgfqpoint{0.833817in}{0.998748in}}%
\pgfusepath{clip}%
\pgfsys@transformshift{2.274190in}{0.472954in}%
\pgftext[left,bottom]{\pgfimage[interpolate=true,width=0.836000in,height=1.000000in]{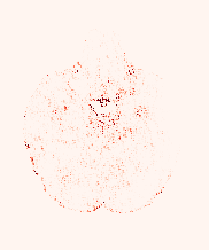}}%
\end{pgfscope}%
\begin{pgfscope}%
\pgfsetrectcap%
\pgfsetmiterjoin%
\pgfsetlinewidth{0.803000pt}%
\definecolor{currentstroke}{rgb}{0.000000,0.000000,0.000000}%
\pgfsetstrokecolor{currentstroke}%
\pgfsetdash{}{0pt}%
\pgfpathmoveto{\pgfqpoint{2.274190in}{0.472954in}}%
\pgfpathlineto{\pgfqpoint{2.274190in}{1.471702in}}%
\pgfusepath{stroke}%
\end{pgfscope}%
\begin{pgfscope}%
\pgfsetrectcap%
\pgfsetmiterjoin%
\pgfsetlinewidth{0.803000pt}%
\definecolor{currentstroke}{rgb}{0.000000,0.000000,0.000000}%
\pgfsetstrokecolor{currentstroke}%
\pgfsetdash{}{0pt}%
\pgfpathmoveto{\pgfqpoint{3.108007in}{0.472954in}}%
\pgfpathlineto{\pgfqpoint{3.108007in}{1.471702in}}%
\pgfusepath{stroke}%
\end{pgfscope}%
\begin{pgfscope}%
\pgfsetrectcap%
\pgfsetmiterjoin%
\pgfsetlinewidth{0.803000pt}%
\definecolor{currentstroke}{rgb}{0.000000,0.000000,0.000000}%
\pgfsetstrokecolor{currentstroke}%
\pgfsetdash{}{0pt}%
\pgfpathmoveto{\pgfqpoint{2.274190in}{0.472954in}}%
\pgfpathlineto{\pgfqpoint{3.108007in}{0.472954in}}%
\pgfusepath{stroke}%
\end{pgfscope}%
\begin{pgfscope}%
\pgfsetrectcap%
\pgfsetmiterjoin%
\pgfsetlinewidth{0.803000pt}%
\definecolor{currentstroke}{rgb}{0.000000,0.000000,0.000000}%
\pgfsetstrokecolor{currentstroke}%
\pgfsetdash{}{0pt}%
\pgfpathmoveto{\pgfqpoint{2.274190in}{1.471702in}}%
\pgfpathlineto{\pgfqpoint{3.108007in}{1.471702in}}%
\pgfusepath{stroke}%
\end{pgfscope}%
\begin{pgfscope}%
\definecolor{textcolor}{rgb}{0.000000,0.000000,0.000000}%
\pgfsetstrokecolor{textcolor}%
\pgfsetfillcolor{textcolor}%
\pgftext[x=2.691099in,y=1.555035in,,base]{\color{textcolor}\rmfamily\fontsize{9.600000}{11.520000}\selectfont Filter 15}%
\end{pgfscope}%
\begin{pgfscope}%
\pgfsetbuttcap%
\pgfsetmiterjoin%
\definecolor{currentfill}{rgb}{1.000000,1.000000,1.000000}%
\pgfsetfillcolor{currentfill}%
\pgfsetlinewidth{0.000000pt}%
\definecolor{currentstroke}{rgb}{0.000000,0.000000,0.000000}%
\pgfsetstrokecolor{currentstroke}%
\pgfsetstrokeopacity{0.000000}%
\pgfsetdash{}{0pt}%
\pgfpathmoveto{\pgfqpoint{3.191389in}{0.472954in}}%
\pgfpathlineto{\pgfqpoint{4.025206in}{0.472954in}}%
\pgfpathlineto{\pgfqpoint{4.025206in}{1.471702in}}%
\pgfpathlineto{\pgfqpoint{3.191389in}{1.471702in}}%
\pgfpathclose%
\pgfusepath{fill}%
\end{pgfscope}%
\begin{pgfscope}%
\pgfpathrectangle{\pgfqpoint{3.191389in}{0.472954in}}{\pgfqpoint{0.833817in}{0.998748in}}%
\pgfusepath{clip}%
\pgfsys@transformshift{3.191389in}{0.472954in}%
\pgftext[left,bottom]{\pgfimage[interpolate=true,width=0.836000in,height=1.000000in]{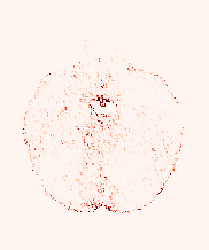}}%
\end{pgfscope}%
\begin{pgfscope}%
\pgfsetrectcap%
\pgfsetmiterjoin%
\pgfsetlinewidth{0.803000pt}%
\definecolor{currentstroke}{rgb}{0.000000,0.000000,0.000000}%
\pgfsetstrokecolor{currentstroke}%
\pgfsetdash{}{0pt}%
\pgfpathmoveto{\pgfqpoint{3.191389in}{0.472954in}}%
\pgfpathlineto{\pgfqpoint{3.191389in}{1.471702in}}%
\pgfusepath{stroke}%
\end{pgfscope}%
\begin{pgfscope}%
\pgfsetrectcap%
\pgfsetmiterjoin%
\pgfsetlinewidth{0.803000pt}%
\definecolor{currentstroke}{rgb}{0.000000,0.000000,0.000000}%
\pgfsetstrokecolor{currentstroke}%
\pgfsetdash{}{0pt}%
\pgfpathmoveto{\pgfqpoint{4.025206in}{0.472954in}}%
\pgfpathlineto{\pgfqpoint{4.025206in}{1.471702in}}%
\pgfusepath{stroke}%
\end{pgfscope}%
\begin{pgfscope}%
\pgfsetrectcap%
\pgfsetmiterjoin%
\pgfsetlinewidth{0.803000pt}%
\definecolor{currentstroke}{rgb}{0.000000,0.000000,0.000000}%
\pgfsetstrokecolor{currentstroke}%
\pgfsetdash{}{0pt}%
\pgfpathmoveto{\pgfqpoint{3.191389in}{0.472954in}}%
\pgfpathlineto{\pgfqpoint{4.025206in}{0.472954in}}%
\pgfusepath{stroke}%
\end{pgfscope}%
\begin{pgfscope}%
\pgfsetrectcap%
\pgfsetmiterjoin%
\pgfsetlinewidth{0.803000pt}%
\definecolor{currentstroke}{rgb}{0.000000,0.000000,0.000000}%
\pgfsetstrokecolor{currentstroke}%
\pgfsetdash{}{0pt}%
\pgfpathmoveto{\pgfqpoint{3.191389in}{1.471702in}}%
\pgfpathlineto{\pgfqpoint{4.025206in}{1.471702in}}%
\pgfusepath{stroke}%
\end{pgfscope}%
\begin{pgfscope}%
\definecolor{textcolor}{rgb}{0.000000,0.000000,0.000000}%
\pgfsetstrokecolor{textcolor}%
\pgfsetfillcolor{textcolor}%
\pgftext[x=3.608298in,y=1.555035in,,base]{\color{textcolor}\rmfamily\fontsize{9.600000}{11.520000}\selectfont Filter 60}%
\end{pgfscope}%
\begin{pgfscope}%
\pgfsetbuttcap%
\pgfsetmiterjoin%
\definecolor{currentfill}{rgb}{1.000000,1.000000,1.000000}%
\pgfsetfillcolor{currentfill}%
\pgfsetlinewidth{0.000000pt}%
\definecolor{currentstroke}{rgb}{0.000000,0.000000,0.000000}%
\pgfsetstrokecolor{currentstroke}%
\pgfsetstrokeopacity{0.000000}%
\pgfsetdash{}{0pt}%
\pgfpathmoveto{\pgfqpoint{4.200308in}{0.635500in}}%
\pgfpathlineto{\pgfqpoint{5.951323in}{0.635500in}}%
\pgfpathlineto{\pgfqpoint{5.951323in}{2.732870in}}%
\pgfpathlineto{\pgfqpoint{4.200308in}{2.732870in}}%
\pgfpathclose%
\pgfusepath{fill}%
\end{pgfscope}%
\begin{pgfscope}%
\pgfpathrectangle{\pgfqpoint{4.200308in}{0.635500in}}{\pgfqpoint{1.751016in}{2.097370in}}%
\pgfusepath{clip}%
\pgfsys@transformshift{4.200308in}{0.635500in}%
\pgftext[left,bottom]{\pgfimage[interpolate=true,width=1.752000in,height=2.100000in]{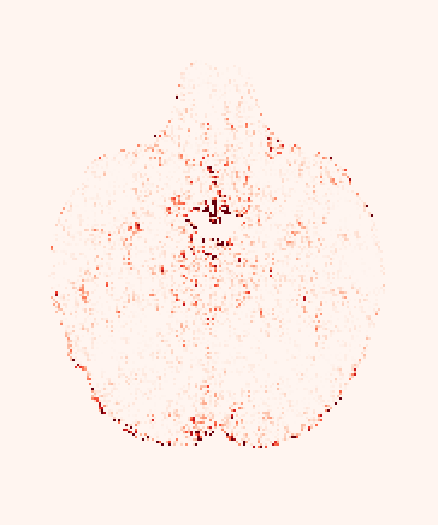}}%
\end{pgfscope}%
\begin{pgfscope}%
\pgfsetrectcap%
\pgfsetmiterjoin%
\pgfsetlinewidth{0.803000pt}%
\definecolor{currentstroke}{rgb}{0.000000,0.000000,0.000000}%
\pgfsetstrokecolor{currentstroke}%
\pgfsetdash{}{0pt}%
\pgfpathmoveto{\pgfqpoint{4.200308in}{0.635500in}}%
\pgfpathlineto{\pgfqpoint{4.200308in}{2.732870in}}%
\pgfusepath{stroke}%
\end{pgfscope}%
\begin{pgfscope}%
\pgfsetrectcap%
\pgfsetmiterjoin%
\pgfsetlinewidth{0.803000pt}%
\definecolor{currentstroke}{rgb}{0.000000,0.000000,0.000000}%
\pgfsetstrokecolor{currentstroke}%
\pgfsetdash{}{0pt}%
\pgfpathmoveto{\pgfqpoint{5.951323in}{0.635500in}}%
\pgfpathlineto{\pgfqpoint{5.951323in}{2.732870in}}%
\pgfusepath{stroke}%
\end{pgfscope}%
\begin{pgfscope}%
\pgfsetrectcap%
\pgfsetmiterjoin%
\pgfsetlinewidth{0.803000pt}%
\definecolor{currentstroke}{rgb}{0.000000,0.000000,0.000000}%
\pgfsetstrokecolor{currentstroke}%
\pgfsetdash{}{0pt}%
\pgfpathmoveto{\pgfqpoint{4.200308in}{0.635500in}}%
\pgfpathlineto{\pgfqpoint{5.951323in}{0.635500in}}%
\pgfusepath{stroke}%
\end{pgfscope}%
\begin{pgfscope}%
\pgfsetrectcap%
\pgfsetmiterjoin%
\pgfsetlinewidth{0.803000pt}%
\definecolor{currentstroke}{rgb}{0.000000,0.000000,0.000000}%
\pgfsetstrokecolor{currentstroke}%
\pgfsetdash{}{0pt}%
\pgfpathmoveto{\pgfqpoint{4.200308in}{2.732870in}}%
\pgfpathlineto{\pgfqpoint{5.951323in}{2.732870in}}%
\pgfusepath{stroke}%
\end{pgfscope}%
\begin{pgfscope}%
\definecolor{textcolor}{rgb}{0.000000,0.000000,0.000000}%
\pgfsetstrokecolor{textcolor}%
\pgfsetfillcolor{textcolor}%
\pgftext[x=5.075815in,y=2.816204in,,base]{\color{textcolor}\rmfamily\fontsize{9.600000}{11.520000}\bfseries\selectfont Score/Final Output}%
\end{pgfscope}%
\begin{pgfscope}%
\definecolor{textcolor}{rgb}{0.000000,0.000000,0.000000}%
\pgfsetstrokecolor{textcolor}%
\pgfsetfillcolor{textcolor}%
\pgftext[x=3.059323in,y=0.235185in,,top]{\color{textcolor}\rmfamily\fontsize{11.000000}{13.200000}\bfseries\selectfont LRP heatmaps of baseline model}%
\end{pgfscope}%
\end{pgfpicture}%
\makeatother%
\endgroup%

%% file: figures/LRP_heatmaps_experiment_PIF-layer_UKB.pgf
\begingroup%
\makeatletter%
\begin{pgfpicture}%
\pgfpathrectangle{\pgfpointorigin}{\pgfqpoint{6.126749in}{2.794969in}}%
\pgfusepath{use as bounding box, clip}%
\begin{pgfscope}%
\pgfsetbuttcap%
\pgfsetmiterjoin%
\definecolor{currentfill}{rgb}{1.000000,1.000000,1.000000}%
\pgfsetfillcolor{currentfill}%
\pgfsetlinewidth{0.000000pt}%
\definecolor{currentstroke}{rgb}{1.000000,1.000000,1.000000}%
\pgfsetstrokecolor{currentstroke}%
\pgfsetdash{}{0pt}%
\pgfpathmoveto{\pgfqpoint{0.000000in}{0.000000in}}%
\pgfpathlineto{\pgfqpoint{6.126749in}{0.000000in}}%
\pgfpathlineto{\pgfqpoint{6.126749in}{2.794969in}}%
\pgfpathlineto{\pgfqpoint{0.000000in}{2.794969in}}%
\pgfpathclose%
\pgfusepath{fill}%
\end{pgfscope}%
\begin{pgfscope}%
\definecolor{textcolor}{rgb}{0.000000,0.000000,0.000000}%
\pgfsetstrokecolor{textcolor}%
\pgfsetfillcolor{textcolor}%
\pgftext[x=0.999614in,y=2.598518in,,base]{\color{textcolor}\rmfamily\fontsize{9.600000}{11.520000}\bfseries\selectfont Patch 10}%
\end{pgfscope}%
\begin{pgfscope}%
\pgfsetbuttcap%
\pgfsetmiterjoin%
\definecolor{currentfill}{rgb}{1.000000,1.000000,1.000000}%
\pgfsetfillcolor{currentfill}%
\pgfsetlinewidth{0.000000pt}%
\definecolor{currentstroke}{rgb}{0.000000,0.000000,0.000000}%
\pgfsetstrokecolor{currentstroke}%
\pgfsetstrokeopacity{0.000000}%
\pgfsetdash{}{0pt}%
\pgfpathmoveto{\pgfqpoint{0.348073in}{1.592308in}}%
\pgfpathlineto{\pgfqpoint{0.968588in}{1.592308in}}%
\pgfpathlineto{\pgfqpoint{0.968588in}{2.335562in}}%
\pgfpathlineto{\pgfqpoint{0.348073in}{2.335562in}}%
\pgfpathclose%
\pgfusepath{fill}%
\end{pgfscope}%
\begin{pgfscope}%
\pgfpathrectangle{\pgfqpoint{0.348073in}{1.592308in}}{\pgfqpoint{0.620515in}{0.743254in}}%
\pgfusepath{clip}%
\pgfsys@transformshift{0.348073in}{1.592308in}%
\pgftext[left,bottom]{\pgfimage[interpolate=true,width=0.624000in,height=0.744000in]{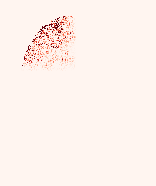}}%
\end{pgfscope}%
\begin{pgfscope}%
\pgfsetrectcap%
\pgfsetmiterjoin%
\pgfsetlinewidth{0.803000pt}%
\definecolor{currentstroke}{rgb}{0.000000,0.000000,0.000000}%
\pgfsetstrokecolor{currentstroke}%
\pgfsetdash{}{0pt}%
\pgfpathmoveto{\pgfqpoint{0.348073in}{1.592308in}}%
\pgfpathlineto{\pgfqpoint{0.348073in}{2.335562in}}%
\pgfusepath{stroke}%
\end{pgfscope}%
\begin{pgfscope}%
\pgfsetrectcap%
\pgfsetmiterjoin%
\pgfsetlinewidth{0.803000pt}%
\definecolor{currentstroke}{rgb}{0.000000,0.000000,0.000000}%
\pgfsetstrokecolor{currentstroke}%
\pgfsetdash{}{0pt}%
\pgfpathmoveto{\pgfqpoint{0.968588in}{1.592308in}}%
\pgfpathlineto{\pgfqpoint{0.968588in}{2.335562in}}%
\pgfusepath{stroke}%
\end{pgfscope}%
\begin{pgfscope}%
\pgfsetrectcap%
\pgfsetmiterjoin%
\pgfsetlinewidth{0.803000pt}%
\definecolor{currentstroke}{rgb}{0.000000,0.000000,0.000000}%
\pgfsetstrokecolor{currentstroke}%
\pgfsetdash{}{0pt}%
\pgfpathmoveto{\pgfqpoint{0.348073in}{1.592308in}}%
\pgfpathlineto{\pgfqpoint{0.968588in}{1.592308in}}%
\pgfusepath{stroke}%
\end{pgfscope}%
\begin{pgfscope}%
\pgfsetrectcap%
\pgfsetmiterjoin%
\pgfsetlinewidth{0.803000pt}%
\definecolor{currentstroke}{rgb}{0.000000,0.000000,0.000000}%
\pgfsetstrokecolor{currentstroke}%
\pgfsetdash{}{0pt}%
\pgfpathmoveto{\pgfqpoint{0.348073in}{2.335562in}}%
\pgfpathlineto{\pgfqpoint{0.968588in}{2.335562in}}%
\pgfusepath{stroke}%
\end{pgfscope}%
\begin{pgfscope}%
\definecolor{textcolor}{rgb}{0.000000,0.000000,0.000000}%
\pgfsetstrokecolor{textcolor}%
\pgfsetfillcolor{textcolor}%
\pgftext[x=0.658331in,y=2.418896in,,base]{\color{textcolor}\rmfamily\fontsize{9.600000}{11.520000}\selectfont Filter 0}%
\end{pgfscope}%
\begin{pgfscope}%
\pgfsetbuttcap%
\pgfsetmiterjoin%
\definecolor{currentfill}{rgb}{1.000000,1.000000,1.000000}%
\pgfsetfillcolor{currentfill}%
\pgfsetlinewidth{0.000000pt}%
\definecolor{currentstroke}{rgb}{0.000000,0.000000,0.000000}%
\pgfsetstrokecolor{currentstroke}%
\pgfsetstrokeopacity{0.000000}%
\pgfsetdash{}{0pt}%
\pgfpathmoveto{\pgfqpoint{1.030640in}{1.592308in}}%
\pgfpathlineto{\pgfqpoint{1.651155in}{1.592308in}}%
\pgfpathlineto{\pgfqpoint{1.651155in}{2.335562in}}%
\pgfpathlineto{\pgfqpoint{1.030640in}{2.335562in}}%
\pgfpathclose%
\pgfusepath{fill}%
\end{pgfscope}%
\begin{pgfscope}%
\pgfpathrectangle{\pgfqpoint{1.030640in}{1.592308in}}{\pgfqpoint{0.620515in}{0.743254in}}%
\pgfusepath{clip}%
\pgfsys@transformshift{1.030640in}{1.592308in}%
\pgftext[left,bottom]{\pgfimage[interpolate=true,width=0.624000in,height=0.744000in]{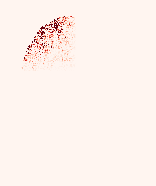}}%
\end{pgfscope}%
\begin{pgfscope}%
\pgfsetrectcap%
\pgfsetmiterjoin%
\pgfsetlinewidth{0.803000pt}%
\definecolor{currentstroke}{rgb}{0.000000,0.000000,0.000000}%
\pgfsetstrokecolor{currentstroke}%
\pgfsetdash{}{0pt}%
\pgfpathmoveto{\pgfqpoint{1.030640in}{1.592308in}}%
\pgfpathlineto{\pgfqpoint{1.030640in}{2.335562in}}%
\pgfusepath{stroke}%
\end{pgfscope}%
\begin{pgfscope}%
\pgfsetrectcap%
\pgfsetmiterjoin%
\pgfsetlinewidth{0.803000pt}%
\definecolor{currentstroke}{rgb}{0.000000,0.000000,0.000000}%
\pgfsetstrokecolor{currentstroke}%
\pgfsetdash{}{0pt}%
\pgfpathmoveto{\pgfqpoint{1.651155in}{1.592308in}}%
\pgfpathlineto{\pgfqpoint{1.651155in}{2.335562in}}%
\pgfusepath{stroke}%
\end{pgfscope}%
\begin{pgfscope}%
\pgfsetrectcap%
\pgfsetmiterjoin%
\pgfsetlinewidth{0.803000pt}%
\definecolor{currentstroke}{rgb}{0.000000,0.000000,0.000000}%
\pgfsetstrokecolor{currentstroke}%
\pgfsetdash{}{0pt}%
\pgfpathmoveto{\pgfqpoint{1.030640in}{1.592308in}}%
\pgfpathlineto{\pgfqpoint{1.651155in}{1.592308in}}%
\pgfusepath{stroke}%
\end{pgfscope}%
\begin{pgfscope}%
\pgfsetrectcap%
\pgfsetmiterjoin%
\pgfsetlinewidth{0.803000pt}%
\definecolor{currentstroke}{rgb}{0.000000,0.000000,0.000000}%
\pgfsetstrokecolor{currentstroke}%
\pgfsetdash{}{0pt}%
\pgfpathmoveto{\pgfqpoint{1.030640in}{2.335562in}}%
\pgfpathlineto{\pgfqpoint{1.651155in}{2.335562in}}%
\pgfusepath{stroke}%
\end{pgfscope}%
\begin{pgfscope}%
\definecolor{textcolor}{rgb}{0.000000,0.000000,0.000000}%
\pgfsetstrokecolor{textcolor}%
\pgfsetfillcolor{textcolor}%
\pgftext[x=1.340897in,y=2.418896in,,base]{\color{textcolor}\rmfamily\fontsize{9.600000}{11.520000}\selectfont Filter 1}%
\end{pgfscope}%
\begin{pgfscope}%
\pgfsetbuttcap%
\pgfsetmiterjoin%
\definecolor{currentfill}{rgb}{1.000000,1.000000,1.000000}%
\pgfsetfillcolor{currentfill}%
\pgfsetlinewidth{0.000000pt}%
\definecolor{currentstroke}{rgb}{0.000000,0.000000,0.000000}%
\pgfsetstrokecolor{currentstroke}%
\pgfsetstrokeopacity{0.000000}%
\pgfsetdash{}{0pt}%
\pgfpathmoveto{\pgfqpoint{0.348073in}{0.489808in}}%
\pgfpathlineto{\pgfqpoint{0.968588in}{0.489808in}}%
\pgfpathlineto{\pgfqpoint{0.968588in}{1.233062in}}%
\pgfpathlineto{\pgfqpoint{0.348073in}{1.233062in}}%
\pgfpathclose%
\pgfusepath{fill}%
\end{pgfscope}%
\begin{pgfscope}%
\pgfpathrectangle{\pgfqpoint{0.348073in}{0.489808in}}{\pgfqpoint{0.620515in}{0.743254in}}%
\pgfusepath{clip}%
\pgfsys@transformshift{0.348073in}{0.489808in}%
\pgftext[left,bottom]{\pgfimage[interpolate=true,width=0.624000in,height=0.744000in]{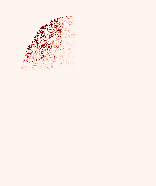}}%
\end{pgfscope}%
\begin{pgfscope}%
\pgfsetrectcap%
\pgfsetmiterjoin%
\pgfsetlinewidth{0.803000pt}%
\definecolor{currentstroke}{rgb}{0.000000,0.000000,0.000000}%
\pgfsetstrokecolor{currentstroke}%
\pgfsetdash{}{0pt}%
\pgfpathmoveto{\pgfqpoint{0.348073in}{0.489808in}}%
\pgfpathlineto{\pgfqpoint{0.348073in}{1.233062in}}%
\pgfusepath{stroke}%
\end{pgfscope}%
\begin{pgfscope}%
\pgfsetrectcap%
\pgfsetmiterjoin%
\pgfsetlinewidth{0.803000pt}%
\definecolor{currentstroke}{rgb}{0.000000,0.000000,0.000000}%
\pgfsetstrokecolor{currentstroke}%
\pgfsetdash{}{0pt}%
\pgfpathmoveto{\pgfqpoint{0.968588in}{0.489808in}}%
\pgfpathlineto{\pgfqpoint{0.968588in}{1.233062in}}%
\pgfusepath{stroke}%
\end{pgfscope}%
\begin{pgfscope}%
\pgfsetrectcap%
\pgfsetmiterjoin%
\pgfsetlinewidth{0.803000pt}%
\definecolor{currentstroke}{rgb}{0.000000,0.000000,0.000000}%
\pgfsetstrokecolor{currentstroke}%
\pgfsetdash{}{0pt}%
\pgfpathmoveto{\pgfqpoint{0.348073in}{0.489808in}}%
\pgfpathlineto{\pgfqpoint{0.968588in}{0.489808in}}%
\pgfusepath{stroke}%
\end{pgfscope}%
\begin{pgfscope}%
\pgfsetrectcap%
\pgfsetmiterjoin%
\pgfsetlinewidth{0.803000pt}%
\definecolor{currentstroke}{rgb}{0.000000,0.000000,0.000000}%
\pgfsetstrokecolor{currentstroke}%
\pgfsetdash{}{0pt}%
\pgfpathmoveto{\pgfqpoint{0.348073in}{1.233062in}}%
\pgfpathlineto{\pgfqpoint{0.968588in}{1.233062in}}%
\pgfusepath{stroke}%
\end{pgfscope}%
\begin{pgfscope}%
\definecolor{textcolor}{rgb}{0.000000,0.000000,0.000000}%
\pgfsetstrokecolor{textcolor}%
\pgfsetfillcolor{textcolor}%
\pgftext[x=0.658331in,y=1.316396in,,base]{\color{textcolor}\rmfamily\fontsize{9.600000}{11.520000}\selectfont Filter 2}%
\end{pgfscope}%
\begin{pgfscope}%
\pgfsetbuttcap%
\pgfsetmiterjoin%
\definecolor{currentfill}{rgb}{1.000000,1.000000,1.000000}%
\pgfsetfillcolor{currentfill}%
\pgfsetlinewidth{0.000000pt}%
\definecolor{currentstroke}{rgb}{0.000000,0.000000,0.000000}%
\pgfsetstrokecolor{currentstroke}%
\pgfsetstrokeopacity{0.000000}%
\pgfsetdash{}{0pt}%
\pgfpathmoveto{\pgfqpoint{1.030640in}{0.489808in}}%
\pgfpathlineto{\pgfqpoint{1.651155in}{0.489808in}}%
\pgfpathlineto{\pgfqpoint{1.651155in}{1.233062in}}%
\pgfpathlineto{\pgfqpoint{1.030640in}{1.233062in}}%
\pgfpathclose%
\pgfusepath{fill}%
\end{pgfscope}%
\begin{pgfscope}%
\pgfpathrectangle{\pgfqpoint{1.030640in}{0.489808in}}{\pgfqpoint{0.620515in}{0.743254in}}%
\pgfusepath{clip}%
\pgfsys@transformshift{1.030640in}{0.489808in}%
\pgftext[left,bottom]{\pgfimage[interpolate=true,width=0.624000in,height=0.744000in]{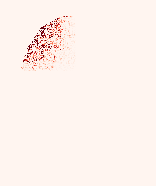}}%
\end{pgfscope}%
\begin{pgfscope}%
\pgfsetrectcap%
\pgfsetmiterjoin%
\pgfsetlinewidth{0.803000pt}%
\definecolor{currentstroke}{rgb}{0.000000,0.000000,0.000000}%
\pgfsetstrokecolor{currentstroke}%
\pgfsetdash{}{0pt}%
\pgfpathmoveto{\pgfqpoint{1.030640in}{0.489808in}}%
\pgfpathlineto{\pgfqpoint{1.030640in}{1.233062in}}%
\pgfusepath{stroke}%
\end{pgfscope}%
\begin{pgfscope}%
\pgfsetrectcap%
\pgfsetmiterjoin%
\pgfsetlinewidth{0.803000pt}%
\definecolor{currentstroke}{rgb}{0.000000,0.000000,0.000000}%
\pgfsetstrokecolor{currentstroke}%
\pgfsetdash{}{0pt}%
\pgfpathmoveto{\pgfqpoint{1.651155in}{0.489808in}}%
\pgfpathlineto{\pgfqpoint{1.651155in}{1.233062in}}%
\pgfusepath{stroke}%
\end{pgfscope}%
\begin{pgfscope}%
\pgfsetrectcap%
\pgfsetmiterjoin%
\pgfsetlinewidth{0.803000pt}%
\definecolor{currentstroke}{rgb}{0.000000,0.000000,0.000000}%
\pgfsetstrokecolor{currentstroke}%
\pgfsetdash{}{0pt}%
\pgfpathmoveto{\pgfqpoint{1.030640in}{0.489808in}}%
\pgfpathlineto{\pgfqpoint{1.651155in}{0.489808in}}%
\pgfusepath{stroke}%
\end{pgfscope}%
\begin{pgfscope}%
\pgfsetrectcap%
\pgfsetmiterjoin%
\pgfsetlinewidth{0.803000pt}%
\definecolor{currentstroke}{rgb}{0.000000,0.000000,0.000000}%
\pgfsetstrokecolor{currentstroke}%
\pgfsetdash{}{0pt}%
\pgfpathmoveto{\pgfqpoint{1.030640in}{1.233062in}}%
\pgfpathlineto{\pgfqpoint{1.651155in}{1.233062in}}%
\pgfusepath{stroke}%
\end{pgfscope}%
\begin{pgfscope}%
\definecolor{textcolor}{rgb}{0.000000,0.000000,0.000000}%
\pgfsetstrokecolor{textcolor}%
\pgfsetfillcolor{textcolor}%
\pgftext[x=1.340897in,y=1.316396in,,base]{\color{textcolor}\rmfamily\fontsize{9.600000}{11.520000}\selectfont Filter 4}%
\end{pgfscope}%
\begin{pgfscope}%
\definecolor{textcolor}{rgb}{0.000000,0.000000,0.000000}%
\pgfsetstrokecolor{textcolor}%
\pgfsetfillcolor{textcolor}%
\pgftext[x=2.433003in,y=2.598518in,,base]{\color{textcolor}\rmfamily\fontsize{9.600000}{11.520000}\bfseries\selectfont Patch 28}%
\end{pgfscope}%
\begin{pgfscope}%
\pgfsetbuttcap%
\pgfsetmiterjoin%
\definecolor{currentfill}{rgb}{1.000000,1.000000,1.000000}%
\pgfsetfillcolor{currentfill}%
\pgfsetlinewidth{0.000000pt}%
\definecolor{currentstroke}{rgb}{0.000000,0.000000,0.000000}%
\pgfsetstrokecolor{currentstroke}%
\pgfsetstrokeopacity{0.000000}%
\pgfsetdash{}{0pt}%
\pgfpathmoveto{\pgfqpoint{1.781463in}{1.592308in}}%
\pgfpathlineto{\pgfqpoint{2.401978in}{1.592308in}}%
\pgfpathlineto{\pgfqpoint{2.401978in}{2.335562in}}%
\pgfpathlineto{\pgfqpoint{1.781463in}{2.335562in}}%
\pgfpathclose%
\pgfusepath{fill}%
\end{pgfscope}%
\begin{pgfscope}%
\pgfpathrectangle{\pgfqpoint{1.781463in}{1.592308in}}{\pgfqpoint{0.620515in}{0.743254in}}%
\pgfusepath{clip}%
\pgfsys@transformshift{1.781463in}{1.592308in}%
\pgftext[left,bottom]{\pgfimage[interpolate=true,width=0.624000in,height=0.744000in]{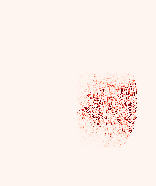}}%
\end{pgfscope}%
\begin{pgfscope}%
\pgfsetrectcap%
\pgfsetmiterjoin%
\pgfsetlinewidth{0.803000pt}%
\definecolor{currentstroke}{rgb}{0.000000,0.000000,0.000000}%
\pgfsetstrokecolor{currentstroke}%
\pgfsetdash{}{0pt}%
\pgfpathmoveto{\pgfqpoint{1.781463in}{1.592308in}}%
\pgfpathlineto{\pgfqpoint{1.781463in}{2.335562in}}%
\pgfusepath{stroke}%
\end{pgfscope}%
\begin{pgfscope}%
\pgfsetrectcap%
\pgfsetmiterjoin%
\pgfsetlinewidth{0.803000pt}%
\definecolor{currentstroke}{rgb}{0.000000,0.000000,0.000000}%
\pgfsetstrokecolor{currentstroke}%
\pgfsetdash{}{0pt}%
\pgfpathmoveto{\pgfqpoint{2.401978in}{1.592308in}}%
\pgfpathlineto{\pgfqpoint{2.401978in}{2.335562in}}%
\pgfusepath{stroke}%
\end{pgfscope}%
\begin{pgfscope}%
\pgfsetrectcap%
\pgfsetmiterjoin%
\pgfsetlinewidth{0.803000pt}%
\definecolor{currentstroke}{rgb}{0.000000,0.000000,0.000000}%
\pgfsetstrokecolor{currentstroke}%
\pgfsetdash{}{0pt}%
\pgfpathmoveto{\pgfqpoint{1.781463in}{1.592308in}}%
\pgfpathlineto{\pgfqpoint{2.401978in}{1.592308in}}%
\pgfusepath{stroke}%
\end{pgfscope}%
\begin{pgfscope}%
\pgfsetrectcap%
\pgfsetmiterjoin%
\pgfsetlinewidth{0.803000pt}%
\definecolor{currentstroke}{rgb}{0.000000,0.000000,0.000000}%
\pgfsetstrokecolor{currentstroke}%
\pgfsetdash{}{0pt}%
\pgfpathmoveto{\pgfqpoint{1.781463in}{2.335562in}}%
\pgfpathlineto{\pgfqpoint{2.401978in}{2.335562in}}%
\pgfusepath{stroke}%
\end{pgfscope}%
\begin{pgfscope}%
\definecolor{textcolor}{rgb}{0.000000,0.000000,0.000000}%
\pgfsetstrokecolor{textcolor}%
\pgfsetfillcolor{textcolor}%
\pgftext[x=2.091720in,y=2.418896in,,base]{\color{textcolor}\rmfamily\fontsize{9.600000}{11.520000}\selectfont Filter 0}%
\end{pgfscope}%
\begin{pgfscope}%
\pgfsetbuttcap%
\pgfsetmiterjoin%
\definecolor{currentfill}{rgb}{1.000000,1.000000,1.000000}%
\pgfsetfillcolor{currentfill}%
\pgfsetlinewidth{0.000000pt}%
\definecolor{currentstroke}{rgb}{0.000000,0.000000,0.000000}%
\pgfsetstrokecolor{currentstroke}%
\pgfsetstrokeopacity{0.000000}%
\pgfsetdash{}{0pt}%
\pgfpathmoveto{\pgfqpoint{2.464029in}{1.592308in}}%
\pgfpathlineto{\pgfqpoint{3.084544in}{1.592308in}}%
\pgfpathlineto{\pgfqpoint{3.084544in}{2.335562in}}%
\pgfpathlineto{\pgfqpoint{2.464029in}{2.335562in}}%
\pgfpathclose%
\pgfusepath{fill}%
\end{pgfscope}%
\begin{pgfscope}%
\pgfpathrectangle{\pgfqpoint{2.464029in}{1.592308in}}{\pgfqpoint{0.620515in}{0.743254in}}%
\pgfusepath{clip}%
\pgfsys@transformshift{2.464029in}{1.592308in}%
\pgftext[left,bottom]{\pgfimage[interpolate=true,width=0.624000in,height=0.744000in]{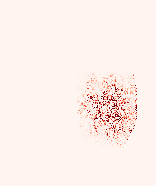}}%
\end{pgfscope}%
\begin{pgfscope}%
\pgfsetrectcap%
\pgfsetmiterjoin%
\pgfsetlinewidth{0.803000pt}%
\definecolor{currentstroke}{rgb}{0.000000,0.000000,0.000000}%
\pgfsetstrokecolor{currentstroke}%
\pgfsetdash{}{0pt}%
\pgfpathmoveto{\pgfqpoint{2.464029in}{1.592308in}}%
\pgfpathlineto{\pgfqpoint{2.464029in}{2.335562in}}%
\pgfusepath{stroke}%
\end{pgfscope}%
\begin{pgfscope}%
\pgfsetrectcap%
\pgfsetmiterjoin%
\pgfsetlinewidth{0.803000pt}%
\definecolor{currentstroke}{rgb}{0.000000,0.000000,0.000000}%
\pgfsetstrokecolor{currentstroke}%
\pgfsetdash{}{0pt}%
\pgfpathmoveto{\pgfqpoint{3.084544in}{1.592308in}}%
\pgfpathlineto{\pgfqpoint{3.084544in}{2.335562in}}%
\pgfusepath{stroke}%
\end{pgfscope}%
\begin{pgfscope}%
\pgfsetrectcap%
\pgfsetmiterjoin%
\pgfsetlinewidth{0.803000pt}%
\definecolor{currentstroke}{rgb}{0.000000,0.000000,0.000000}%
\pgfsetstrokecolor{currentstroke}%
\pgfsetdash{}{0pt}%
\pgfpathmoveto{\pgfqpoint{2.464029in}{1.592308in}}%
\pgfpathlineto{\pgfqpoint{3.084544in}{1.592308in}}%
\pgfusepath{stroke}%
\end{pgfscope}%
\begin{pgfscope}%
\pgfsetrectcap%
\pgfsetmiterjoin%
\pgfsetlinewidth{0.803000pt}%
\definecolor{currentstroke}{rgb}{0.000000,0.000000,0.000000}%
\pgfsetstrokecolor{currentstroke}%
\pgfsetdash{}{0pt}%
\pgfpathmoveto{\pgfqpoint{2.464029in}{2.335562in}}%
\pgfpathlineto{\pgfqpoint{3.084544in}{2.335562in}}%
\pgfusepath{stroke}%
\end{pgfscope}%
\begin{pgfscope}%
\definecolor{textcolor}{rgb}{0.000000,0.000000,0.000000}%
\pgfsetstrokecolor{textcolor}%
\pgfsetfillcolor{textcolor}%
\pgftext[x=2.774287in,y=2.418896in,,base]{\color{textcolor}\rmfamily\fontsize{9.600000}{11.520000}\selectfont Filter 1}%
\end{pgfscope}%
\begin{pgfscope}%
\pgfsetbuttcap%
\pgfsetmiterjoin%
\definecolor{currentfill}{rgb}{1.000000,1.000000,1.000000}%
\pgfsetfillcolor{currentfill}%
\pgfsetlinewidth{0.000000pt}%
\definecolor{currentstroke}{rgb}{0.000000,0.000000,0.000000}%
\pgfsetstrokecolor{currentstroke}%
\pgfsetstrokeopacity{0.000000}%
\pgfsetdash{}{0pt}%
\pgfpathmoveto{\pgfqpoint{1.781463in}{0.489808in}}%
\pgfpathlineto{\pgfqpoint{2.401978in}{0.489808in}}%
\pgfpathlineto{\pgfqpoint{2.401978in}{1.233062in}}%
\pgfpathlineto{\pgfqpoint{1.781463in}{1.233062in}}%
\pgfpathclose%
\pgfusepath{fill}%
\end{pgfscope}%
\begin{pgfscope}%
\pgfpathrectangle{\pgfqpoint{1.781463in}{0.489808in}}{\pgfqpoint{0.620515in}{0.743254in}}%
\pgfusepath{clip}%
\pgfsys@transformshift{1.781463in}{0.489808in}%
\pgftext[left,bottom]{\pgfimage[interpolate=true,width=0.624000in,height=0.744000in]{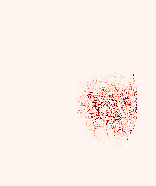}}%
\end{pgfscope}%
\begin{pgfscope}%
\pgfsetrectcap%
\pgfsetmiterjoin%
\pgfsetlinewidth{0.803000pt}%
\definecolor{currentstroke}{rgb}{0.000000,0.000000,0.000000}%
\pgfsetstrokecolor{currentstroke}%
\pgfsetdash{}{0pt}%
\pgfpathmoveto{\pgfqpoint{1.781463in}{0.489808in}}%
\pgfpathlineto{\pgfqpoint{1.781463in}{1.233062in}}%
\pgfusepath{stroke}%
\end{pgfscope}%
\begin{pgfscope}%
\pgfsetrectcap%
\pgfsetmiterjoin%
\pgfsetlinewidth{0.803000pt}%
\definecolor{currentstroke}{rgb}{0.000000,0.000000,0.000000}%
\pgfsetstrokecolor{currentstroke}%
\pgfsetdash{}{0pt}%
\pgfpathmoveto{\pgfqpoint{2.401978in}{0.489808in}}%
\pgfpathlineto{\pgfqpoint{2.401978in}{1.233062in}}%
\pgfusepath{stroke}%
\end{pgfscope}%
\begin{pgfscope}%
\pgfsetrectcap%
\pgfsetmiterjoin%
\pgfsetlinewidth{0.803000pt}%
\definecolor{currentstroke}{rgb}{0.000000,0.000000,0.000000}%
\pgfsetstrokecolor{currentstroke}%
\pgfsetdash{}{0pt}%
\pgfpathmoveto{\pgfqpoint{1.781463in}{0.489808in}}%
\pgfpathlineto{\pgfqpoint{2.401978in}{0.489808in}}%
\pgfusepath{stroke}%
\end{pgfscope}%
\begin{pgfscope}%
\pgfsetrectcap%
\pgfsetmiterjoin%
\pgfsetlinewidth{0.803000pt}%
\definecolor{currentstroke}{rgb}{0.000000,0.000000,0.000000}%
\pgfsetstrokecolor{currentstroke}%
\pgfsetdash{}{0pt}%
\pgfpathmoveto{\pgfqpoint{1.781463in}{1.233062in}}%
\pgfpathlineto{\pgfqpoint{2.401978in}{1.233062in}}%
\pgfusepath{stroke}%
\end{pgfscope}%
\begin{pgfscope}%
\definecolor{textcolor}{rgb}{0.000000,0.000000,0.000000}%
\pgfsetstrokecolor{textcolor}%
\pgfsetfillcolor{textcolor}%
\pgftext[x=2.091720in,y=1.316396in,,base]{\color{textcolor}\rmfamily\fontsize{9.600000}{11.520000}\selectfont Filter 2}%
\end{pgfscope}%
\begin{pgfscope}%
\pgfsetbuttcap%
\pgfsetmiterjoin%
\definecolor{currentfill}{rgb}{1.000000,1.000000,1.000000}%
\pgfsetfillcolor{currentfill}%
\pgfsetlinewidth{0.000000pt}%
\definecolor{currentstroke}{rgb}{0.000000,0.000000,0.000000}%
\pgfsetstrokecolor{currentstroke}%
\pgfsetstrokeopacity{0.000000}%
\pgfsetdash{}{0pt}%
\pgfpathmoveto{\pgfqpoint{2.464029in}{0.489808in}}%
\pgfpathlineto{\pgfqpoint{3.084544in}{0.489808in}}%
\pgfpathlineto{\pgfqpoint{3.084544in}{1.233062in}}%
\pgfpathlineto{\pgfqpoint{2.464029in}{1.233062in}}%
\pgfpathclose%
\pgfusepath{fill}%
\end{pgfscope}%
\begin{pgfscope}%
\pgfpathrectangle{\pgfqpoint{2.464029in}{0.489808in}}{\pgfqpoint{0.620515in}{0.743254in}}%
\pgfusepath{clip}%
\pgfsys@transformshift{2.464029in}{0.489808in}%
\pgftext[left,bottom]{\pgfimage[interpolate=true,width=0.624000in,height=0.744000in]{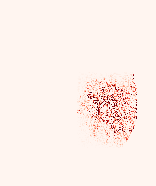}}%
\end{pgfscope}%
\begin{pgfscope}%
\pgfsetrectcap%
\pgfsetmiterjoin%
\pgfsetlinewidth{0.803000pt}%
\definecolor{currentstroke}{rgb}{0.000000,0.000000,0.000000}%
\pgfsetstrokecolor{currentstroke}%
\pgfsetdash{}{0pt}%
\pgfpathmoveto{\pgfqpoint{2.464029in}{0.489808in}}%
\pgfpathlineto{\pgfqpoint{2.464029in}{1.233062in}}%
\pgfusepath{stroke}%
\end{pgfscope}%
\begin{pgfscope}%
\pgfsetrectcap%
\pgfsetmiterjoin%
\pgfsetlinewidth{0.803000pt}%
\definecolor{currentstroke}{rgb}{0.000000,0.000000,0.000000}%
\pgfsetstrokecolor{currentstroke}%
\pgfsetdash{}{0pt}%
\pgfpathmoveto{\pgfqpoint{3.084544in}{0.489808in}}%
\pgfpathlineto{\pgfqpoint{3.084544in}{1.233062in}}%
\pgfusepath{stroke}%
\end{pgfscope}%
\begin{pgfscope}%
\pgfsetrectcap%
\pgfsetmiterjoin%
\pgfsetlinewidth{0.803000pt}%
\definecolor{currentstroke}{rgb}{0.000000,0.000000,0.000000}%
\pgfsetstrokecolor{currentstroke}%
\pgfsetdash{}{0pt}%
\pgfpathmoveto{\pgfqpoint{2.464029in}{0.489808in}}%
\pgfpathlineto{\pgfqpoint{3.084544in}{0.489808in}}%
\pgfusepath{stroke}%
\end{pgfscope}%
\begin{pgfscope}%
\pgfsetrectcap%
\pgfsetmiterjoin%
\pgfsetlinewidth{0.803000pt}%
\definecolor{currentstroke}{rgb}{0.000000,0.000000,0.000000}%
\pgfsetstrokecolor{currentstroke}%
\pgfsetdash{}{0pt}%
\pgfpathmoveto{\pgfqpoint{2.464029in}{1.233062in}}%
\pgfpathlineto{\pgfqpoint{3.084544in}{1.233062in}}%
\pgfusepath{stroke}%
\end{pgfscope}%
\begin{pgfscope}%
\definecolor{textcolor}{rgb}{0.000000,0.000000,0.000000}%
\pgfsetstrokecolor{textcolor}%
\pgfsetfillcolor{textcolor}%
\pgftext[x=2.774287in,y=1.316396in,,base]{\color{textcolor}\rmfamily\fontsize{9.600000}{11.520000}\selectfont Filter 4}%
\end{pgfscope}%
\begin{pgfscope}%
\definecolor{textcolor}{rgb}{0.000000,0.000000,0.000000}%
\pgfsetstrokecolor{textcolor}%
\pgfsetfillcolor{textcolor}%
\pgftext[x=3.866393in,y=2.598518in,,base]{\color{textcolor}\rmfamily\fontsize{9.600000}{11.520000}\bfseries\selectfont Patch 40}%
\end{pgfscope}%
\begin{pgfscope}%
\pgfsetbuttcap%
\pgfsetmiterjoin%
\definecolor{currentfill}{rgb}{1.000000,1.000000,1.000000}%
\pgfsetfillcolor{currentfill}%
\pgfsetlinewidth{0.000000pt}%
\definecolor{currentstroke}{rgb}{0.000000,0.000000,0.000000}%
\pgfsetstrokecolor{currentstroke}%
\pgfsetstrokeopacity{0.000000}%
\pgfsetdash{}{0pt}%
\pgfpathmoveto{\pgfqpoint{3.214852in}{1.592308in}}%
\pgfpathlineto{\pgfqpoint{3.835367in}{1.592308in}}%
\pgfpathlineto{\pgfqpoint{3.835367in}{2.335562in}}%
\pgfpathlineto{\pgfqpoint{3.214852in}{2.335562in}}%
\pgfpathclose%
\pgfusepath{fill}%
\end{pgfscope}%
\begin{pgfscope}%
\pgfpathrectangle{\pgfqpoint{3.214852in}{1.592308in}}{\pgfqpoint{0.620515in}{0.743254in}}%
\pgfusepath{clip}%
\pgfsys@transformshift{3.214852in}{1.592308in}%
\pgftext[left,bottom]{\pgfimage[interpolate=true,width=0.624000in,height=0.744000in]{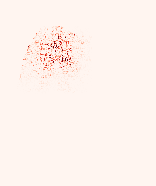}}%
\end{pgfscope}%
\begin{pgfscope}%
\pgfsetrectcap%
\pgfsetmiterjoin%
\pgfsetlinewidth{0.803000pt}%
\definecolor{currentstroke}{rgb}{0.000000,0.000000,0.000000}%
\pgfsetstrokecolor{currentstroke}%
\pgfsetdash{}{0pt}%
\pgfpathmoveto{\pgfqpoint{3.214852in}{1.592308in}}%
\pgfpathlineto{\pgfqpoint{3.214852in}{2.335562in}}%
\pgfusepath{stroke}%
\end{pgfscope}%
\begin{pgfscope}%
\pgfsetrectcap%
\pgfsetmiterjoin%
\pgfsetlinewidth{0.803000pt}%
\definecolor{currentstroke}{rgb}{0.000000,0.000000,0.000000}%
\pgfsetstrokecolor{currentstroke}%
\pgfsetdash{}{0pt}%
\pgfpathmoveto{\pgfqpoint{3.835367in}{1.592308in}}%
\pgfpathlineto{\pgfqpoint{3.835367in}{2.335562in}}%
\pgfusepath{stroke}%
\end{pgfscope}%
\begin{pgfscope}%
\pgfsetrectcap%
\pgfsetmiterjoin%
\pgfsetlinewidth{0.803000pt}%
\definecolor{currentstroke}{rgb}{0.000000,0.000000,0.000000}%
\pgfsetstrokecolor{currentstroke}%
\pgfsetdash{}{0pt}%
\pgfpathmoveto{\pgfqpoint{3.214852in}{1.592308in}}%
\pgfpathlineto{\pgfqpoint{3.835367in}{1.592308in}}%
\pgfusepath{stroke}%
\end{pgfscope}%
\begin{pgfscope}%
\pgfsetrectcap%
\pgfsetmiterjoin%
\pgfsetlinewidth{0.803000pt}%
\definecolor{currentstroke}{rgb}{0.000000,0.000000,0.000000}%
\pgfsetstrokecolor{currentstroke}%
\pgfsetdash{}{0pt}%
\pgfpathmoveto{\pgfqpoint{3.214852in}{2.335562in}}%
\pgfpathlineto{\pgfqpoint{3.835367in}{2.335562in}}%
\pgfusepath{stroke}%
\end{pgfscope}%
\begin{pgfscope}%
\definecolor{textcolor}{rgb}{0.000000,0.000000,0.000000}%
\pgfsetstrokecolor{textcolor}%
\pgfsetfillcolor{textcolor}%
\pgftext[x=3.525110in,y=2.418896in,,base]{\color{textcolor}\rmfamily\fontsize{9.600000}{11.520000}\selectfont Filter 0}%
\end{pgfscope}%
\begin{pgfscope}%
\pgfsetbuttcap%
\pgfsetmiterjoin%
\definecolor{currentfill}{rgb}{1.000000,1.000000,1.000000}%
\pgfsetfillcolor{currentfill}%
\pgfsetlinewidth{0.000000pt}%
\definecolor{currentstroke}{rgb}{0.000000,0.000000,0.000000}%
\pgfsetstrokecolor{currentstroke}%
\pgfsetstrokeopacity{0.000000}%
\pgfsetdash{}{0pt}%
\pgfpathmoveto{\pgfqpoint{3.897419in}{1.592308in}}%
\pgfpathlineto{\pgfqpoint{4.517934in}{1.592308in}}%
\pgfpathlineto{\pgfqpoint{4.517934in}{2.335562in}}%
\pgfpathlineto{\pgfqpoint{3.897419in}{2.335562in}}%
\pgfpathclose%
\pgfusepath{fill}%
\end{pgfscope}%
\begin{pgfscope}%
\pgfpathrectangle{\pgfqpoint{3.897419in}{1.592308in}}{\pgfqpoint{0.620515in}{0.743254in}}%
\pgfusepath{clip}%
\pgfsys@transformshift{3.897419in}{1.592308in}%
\pgftext[left,bottom]{\pgfimage[interpolate=true,width=0.624000in,height=0.744000in]{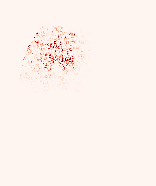}}%
\end{pgfscope}%
\begin{pgfscope}%
\pgfsetrectcap%
\pgfsetmiterjoin%
\pgfsetlinewidth{0.803000pt}%
\definecolor{currentstroke}{rgb}{0.000000,0.000000,0.000000}%
\pgfsetstrokecolor{currentstroke}%
\pgfsetdash{}{0pt}%
\pgfpathmoveto{\pgfqpoint{3.897419in}{1.592308in}}%
\pgfpathlineto{\pgfqpoint{3.897419in}{2.335562in}}%
\pgfusepath{stroke}%
\end{pgfscope}%
\begin{pgfscope}%
\pgfsetrectcap%
\pgfsetmiterjoin%
\pgfsetlinewidth{0.803000pt}%
\definecolor{currentstroke}{rgb}{0.000000,0.000000,0.000000}%
\pgfsetstrokecolor{currentstroke}%
\pgfsetdash{}{0pt}%
\pgfpathmoveto{\pgfqpoint{4.517934in}{1.592308in}}%
\pgfpathlineto{\pgfqpoint{4.517934in}{2.335562in}}%
\pgfusepath{stroke}%
\end{pgfscope}%
\begin{pgfscope}%
\pgfsetrectcap%
\pgfsetmiterjoin%
\pgfsetlinewidth{0.803000pt}%
\definecolor{currentstroke}{rgb}{0.000000,0.000000,0.000000}%
\pgfsetstrokecolor{currentstroke}%
\pgfsetdash{}{0pt}%
\pgfpathmoveto{\pgfqpoint{3.897419in}{1.592308in}}%
\pgfpathlineto{\pgfqpoint{4.517934in}{1.592308in}}%
\pgfusepath{stroke}%
\end{pgfscope}%
\begin{pgfscope}%
\pgfsetrectcap%
\pgfsetmiterjoin%
\pgfsetlinewidth{0.803000pt}%
\definecolor{currentstroke}{rgb}{0.000000,0.000000,0.000000}%
\pgfsetstrokecolor{currentstroke}%
\pgfsetdash{}{0pt}%
\pgfpathmoveto{\pgfqpoint{3.897419in}{2.335562in}}%
\pgfpathlineto{\pgfqpoint{4.517934in}{2.335562in}}%
\pgfusepath{stroke}%
\end{pgfscope}%
\begin{pgfscope}%
\definecolor{textcolor}{rgb}{0.000000,0.000000,0.000000}%
\pgfsetstrokecolor{textcolor}%
\pgfsetfillcolor{textcolor}%
\pgftext[x=4.207676in,y=2.418896in,,base]{\color{textcolor}\rmfamily\fontsize{9.600000}{11.520000}\selectfont Filter 1}%
\end{pgfscope}%
\begin{pgfscope}%
\pgfsetbuttcap%
\pgfsetmiterjoin%
\definecolor{currentfill}{rgb}{1.000000,1.000000,1.000000}%
\pgfsetfillcolor{currentfill}%
\pgfsetlinewidth{0.000000pt}%
\definecolor{currentstroke}{rgb}{0.000000,0.000000,0.000000}%
\pgfsetstrokecolor{currentstroke}%
\pgfsetstrokeopacity{0.000000}%
\pgfsetdash{}{0pt}%
\pgfpathmoveto{\pgfqpoint{3.214852in}{0.489808in}}%
\pgfpathlineto{\pgfqpoint{3.835367in}{0.489808in}}%
\pgfpathlineto{\pgfqpoint{3.835367in}{1.233062in}}%
\pgfpathlineto{\pgfqpoint{3.214852in}{1.233062in}}%
\pgfpathclose%
\pgfusepath{fill}%
\end{pgfscope}%
\begin{pgfscope}%
\pgfpathrectangle{\pgfqpoint{3.214852in}{0.489808in}}{\pgfqpoint{0.620515in}{0.743254in}}%
\pgfusepath{clip}%
\pgfsys@transformshift{3.214852in}{0.489808in}%
\pgftext[left,bottom]{\pgfimage[interpolate=true,width=0.624000in,height=0.744000in]{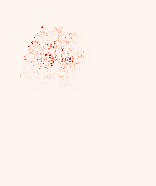}}%
\end{pgfscope}%
\begin{pgfscope}%
\pgfsetrectcap%
\pgfsetmiterjoin%
\pgfsetlinewidth{0.803000pt}%
\definecolor{currentstroke}{rgb}{0.000000,0.000000,0.000000}%
\pgfsetstrokecolor{currentstroke}%
\pgfsetdash{}{0pt}%
\pgfpathmoveto{\pgfqpoint{3.214852in}{0.489808in}}%
\pgfpathlineto{\pgfqpoint{3.214852in}{1.233062in}}%
\pgfusepath{stroke}%
\end{pgfscope}%
\begin{pgfscope}%
\pgfsetrectcap%
\pgfsetmiterjoin%
\pgfsetlinewidth{0.803000pt}%
\definecolor{currentstroke}{rgb}{0.000000,0.000000,0.000000}%
\pgfsetstrokecolor{currentstroke}%
\pgfsetdash{}{0pt}%
\pgfpathmoveto{\pgfqpoint{3.835367in}{0.489808in}}%
\pgfpathlineto{\pgfqpoint{3.835367in}{1.233062in}}%
\pgfusepath{stroke}%
\end{pgfscope}%
\begin{pgfscope}%
\pgfsetrectcap%
\pgfsetmiterjoin%
\pgfsetlinewidth{0.803000pt}%
\definecolor{currentstroke}{rgb}{0.000000,0.000000,0.000000}%
\pgfsetstrokecolor{currentstroke}%
\pgfsetdash{}{0pt}%
\pgfpathmoveto{\pgfqpoint{3.214852in}{0.489808in}}%
\pgfpathlineto{\pgfqpoint{3.835367in}{0.489808in}}%
\pgfusepath{stroke}%
\end{pgfscope}%
\begin{pgfscope}%
\pgfsetrectcap%
\pgfsetmiterjoin%
\pgfsetlinewidth{0.803000pt}%
\definecolor{currentstroke}{rgb}{0.000000,0.000000,0.000000}%
\pgfsetstrokecolor{currentstroke}%
\pgfsetdash{}{0pt}%
\pgfpathmoveto{\pgfqpoint{3.214852in}{1.233062in}}%
\pgfpathlineto{\pgfqpoint{3.835367in}{1.233062in}}%
\pgfusepath{stroke}%
\end{pgfscope}%
\begin{pgfscope}%
\definecolor{textcolor}{rgb}{0.000000,0.000000,0.000000}%
\pgfsetstrokecolor{textcolor}%
\pgfsetfillcolor{textcolor}%
\pgftext[x=3.525110in,y=1.316396in,,base]{\color{textcolor}\rmfamily\fontsize{9.600000}{11.520000}\selectfont Filter 2}%
\end{pgfscope}%
\begin{pgfscope}%
\pgfsetbuttcap%
\pgfsetmiterjoin%
\definecolor{currentfill}{rgb}{1.000000,1.000000,1.000000}%
\pgfsetfillcolor{currentfill}%
\pgfsetlinewidth{0.000000pt}%
\definecolor{currentstroke}{rgb}{0.000000,0.000000,0.000000}%
\pgfsetstrokecolor{currentstroke}%
\pgfsetstrokeopacity{0.000000}%
\pgfsetdash{}{0pt}%
\pgfpathmoveto{\pgfqpoint{3.897419in}{0.489808in}}%
\pgfpathlineto{\pgfqpoint{4.517934in}{0.489808in}}%
\pgfpathlineto{\pgfqpoint{4.517934in}{1.233062in}}%
\pgfpathlineto{\pgfqpoint{3.897419in}{1.233062in}}%
\pgfpathclose%
\pgfusepath{fill}%
\end{pgfscope}%
\begin{pgfscope}%
\pgfpathrectangle{\pgfqpoint{3.897419in}{0.489808in}}{\pgfqpoint{0.620515in}{0.743254in}}%
\pgfusepath{clip}%
\pgfsys@transformshift{3.897419in}{0.489808in}%
\pgftext[left,bottom]{\pgfimage[interpolate=true,width=0.624000in,height=0.744000in]{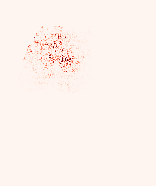}}%
\end{pgfscope}%
\begin{pgfscope}%
\pgfsetrectcap%
\pgfsetmiterjoin%
\pgfsetlinewidth{0.803000pt}%
\definecolor{currentstroke}{rgb}{0.000000,0.000000,0.000000}%
\pgfsetstrokecolor{currentstroke}%
\pgfsetdash{}{0pt}%
\pgfpathmoveto{\pgfqpoint{3.897419in}{0.489808in}}%
\pgfpathlineto{\pgfqpoint{3.897419in}{1.233062in}}%
\pgfusepath{stroke}%
\end{pgfscope}%
\begin{pgfscope}%
\pgfsetrectcap%
\pgfsetmiterjoin%
\pgfsetlinewidth{0.803000pt}%
\definecolor{currentstroke}{rgb}{0.000000,0.000000,0.000000}%
\pgfsetstrokecolor{currentstroke}%
\pgfsetdash{}{0pt}%
\pgfpathmoveto{\pgfqpoint{4.517934in}{0.489808in}}%
\pgfpathlineto{\pgfqpoint{4.517934in}{1.233062in}}%
\pgfusepath{stroke}%
\end{pgfscope}%
\begin{pgfscope}%
\pgfsetrectcap%
\pgfsetmiterjoin%
\pgfsetlinewidth{0.803000pt}%
\definecolor{currentstroke}{rgb}{0.000000,0.000000,0.000000}%
\pgfsetstrokecolor{currentstroke}%
\pgfsetdash{}{0pt}%
\pgfpathmoveto{\pgfqpoint{3.897419in}{0.489808in}}%
\pgfpathlineto{\pgfqpoint{4.517934in}{0.489808in}}%
\pgfusepath{stroke}%
\end{pgfscope}%
\begin{pgfscope}%
\pgfsetrectcap%
\pgfsetmiterjoin%
\pgfsetlinewidth{0.803000pt}%
\definecolor{currentstroke}{rgb}{0.000000,0.000000,0.000000}%
\pgfsetstrokecolor{currentstroke}%
\pgfsetdash{}{0pt}%
\pgfpathmoveto{\pgfqpoint{3.897419in}{1.233062in}}%
\pgfpathlineto{\pgfqpoint{4.517934in}{1.233062in}}%
\pgfusepath{stroke}%
\end{pgfscope}%
\begin{pgfscope}%
\definecolor{textcolor}{rgb}{0.000000,0.000000,0.000000}%
\pgfsetstrokecolor{textcolor}%
\pgfsetfillcolor{textcolor}%
\pgftext[x=4.207676in,y=1.316396in,,base]{\color{textcolor}\rmfamily\fontsize{9.600000}{11.520000}\selectfont Filter 4}%
\end{pgfscope}%
\begin{pgfscope}%
\definecolor{textcolor}{rgb}{0.000000,0.000000,0.000000}%
\pgfsetstrokecolor{textcolor}%
\pgfsetfillcolor{textcolor}%
\pgftext[x=5.299782in,y=2.598518in,,base]{\color{textcolor}\rmfamily\fontsize{9.600000}{11.520000}\bfseries\selectfont Patch 41}%
\end{pgfscope}%
\begin{pgfscope}%
\pgfsetbuttcap%
\pgfsetmiterjoin%
\definecolor{currentfill}{rgb}{1.000000,1.000000,1.000000}%
\pgfsetfillcolor{currentfill}%
\pgfsetlinewidth{0.000000pt}%
\definecolor{currentstroke}{rgb}{0.000000,0.000000,0.000000}%
\pgfsetstrokecolor{currentstroke}%
\pgfsetstrokeopacity{0.000000}%
\pgfsetdash{}{0pt}%
\pgfpathmoveto{\pgfqpoint{4.648242in}{1.592308in}}%
\pgfpathlineto{\pgfqpoint{5.268757in}{1.592308in}}%
\pgfpathlineto{\pgfqpoint{5.268757in}{2.335562in}}%
\pgfpathlineto{\pgfqpoint{4.648242in}{2.335562in}}%
\pgfpathclose%
\pgfusepath{fill}%
\end{pgfscope}%
\begin{pgfscope}%
\pgfpathrectangle{\pgfqpoint{4.648242in}{1.592308in}}{\pgfqpoint{0.620515in}{0.743254in}}%
\pgfusepath{clip}%
\pgfsys@transformshift{4.648242in}{1.592308in}%
\pgftext[left,bottom]{\pgfimage[interpolate=true,width=0.624000in,height=0.744000in]{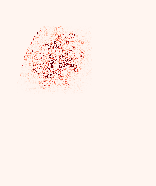}}%
\end{pgfscope}%
\begin{pgfscope}%
\pgfsetrectcap%
\pgfsetmiterjoin%
\pgfsetlinewidth{0.803000pt}%
\definecolor{currentstroke}{rgb}{0.000000,0.000000,0.000000}%
\pgfsetstrokecolor{currentstroke}%
\pgfsetdash{}{0pt}%
\pgfpathmoveto{\pgfqpoint{4.648242in}{1.592308in}}%
\pgfpathlineto{\pgfqpoint{4.648242in}{2.335562in}}%
\pgfusepath{stroke}%
\end{pgfscope}%
\begin{pgfscope}%
\pgfsetrectcap%
\pgfsetmiterjoin%
\pgfsetlinewidth{0.803000pt}%
\definecolor{currentstroke}{rgb}{0.000000,0.000000,0.000000}%
\pgfsetstrokecolor{currentstroke}%
\pgfsetdash{}{0pt}%
\pgfpathmoveto{\pgfqpoint{5.268757in}{1.592308in}}%
\pgfpathlineto{\pgfqpoint{5.268757in}{2.335562in}}%
\pgfusepath{stroke}%
\end{pgfscope}%
\begin{pgfscope}%
\pgfsetrectcap%
\pgfsetmiterjoin%
\pgfsetlinewidth{0.803000pt}%
\definecolor{currentstroke}{rgb}{0.000000,0.000000,0.000000}%
\pgfsetstrokecolor{currentstroke}%
\pgfsetdash{}{0pt}%
\pgfpathmoveto{\pgfqpoint{4.648242in}{1.592308in}}%
\pgfpathlineto{\pgfqpoint{5.268757in}{1.592308in}}%
\pgfusepath{stroke}%
\end{pgfscope}%
\begin{pgfscope}%
\pgfsetrectcap%
\pgfsetmiterjoin%
\pgfsetlinewidth{0.803000pt}%
\definecolor{currentstroke}{rgb}{0.000000,0.000000,0.000000}%
\pgfsetstrokecolor{currentstroke}%
\pgfsetdash{}{0pt}%
\pgfpathmoveto{\pgfqpoint{4.648242in}{2.335562in}}%
\pgfpathlineto{\pgfqpoint{5.268757in}{2.335562in}}%
\pgfusepath{stroke}%
\end{pgfscope}%
\begin{pgfscope}%
\definecolor{textcolor}{rgb}{0.000000,0.000000,0.000000}%
\pgfsetstrokecolor{textcolor}%
\pgfsetfillcolor{textcolor}%
\pgftext[x=4.958499in,y=2.418896in,,base]{\color{textcolor}\rmfamily\fontsize{9.600000}{11.520000}\selectfont Filter 0}%
\end{pgfscope}%
\begin{pgfscope}%
\pgfsetbuttcap%
\pgfsetmiterjoin%
\definecolor{currentfill}{rgb}{1.000000,1.000000,1.000000}%
\pgfsetfillcolor{currentfill}%
\pgfsetlinewidth{0.000000pt}%
\definecolor{currentstroke}{rgb}{0.000000,0.000000,0.000000}%
\pgfsetstrokecolor{currentstroke}%
\pgfsetstrokeopacity{0.000000}%
\pgfsetdash{}{0pt}%
\pgfpathmoveto{\pgfqpoint{5.330808in}{1.592308in}}%
\pgfpathlineto{\pgfqpoint{5.951323in}{1.592308in}}%
\pgfpathlineto{\pgfqpoint{5.951323in}{2.335562in}}%
\pgfpathlineto{\pgfqpoint{5.330808in}{2.335562in}}%
\pgfpathclose%
\pgfusepath{fill}%
\end{pgfscope}%
\begin{pgfscope}%
\pgfpathrectangle{\pgfqpoint{5.330808in}{1.592308in}}{\pgfqpoint{0.620515in}{0.743254in}}%
\pgfusepath{clip}%
\pgfsys@transformshift{5.330808in}{1.592308in}%
\pgftext[left,bottom]{\pgfimage[interpolate=true,width=0.624000in,height=0.744000in]{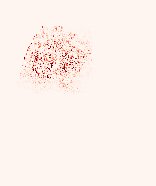}}%
\end{pgfscope}%
\begin{pgfscope}%
\pgfsetrectcap%
\pgfsetmiterjoin%
\pgfsetlinewidth{0.803000pt}%
\definecolor{currentstroke}{rgb}{0.000000,0.000000,0.000000}%
\pgfsetstrokecolor{currentstroke}%
\pgfsetdash{}{0pt}%
\pgfpathmoveto{\pgfqpoint{5.330808in}{1.592308in}}%
\pgfpathlineto{\pgfqpoint{5.330808in}{2.335562in}}%
\pgfusepath{stroke}%
\end{pgfscope}%
\begin{pgfscope}%
\pgfsetrectcap%
\pgfsetmiterjoin%
\pgfsetlinewidth{0.803000pt}%
\definecolor{currentstroke}{rgb}{0.000000,0.000000,0.000000}%
\pgfsetstrokecolor{currentstroke}%
\pgfsetdash{}{0pt}%
\pgfpathmoveto{\pgfqpoint{5.951323in}{1.592308in}}%
\pgfpathlineto{\pgfqpoint{5.951323in}{2.335562in}}%
\pgfusepath{stroke}%
\end{pgfscope}%
\begin{pgfscope}%
\pgfsetrectcap%
\pgfsetmiterjoin%
\pgfsetlinewidth{0.803000pt}%
\definecolor{currentstroke}{rgb}{0.000000,0.000000,0.000000}%
\pgfsetstrokecolor{currentstroke}%
\pgfsetdash{}{0pt}%
\pgfpathmoveto{\pgfqpoint{5.330808in}{1.592308in}}%
\pgfpathlineto{\pgfqpoint{5.951323in}{1.592308in}}%
\pgfusepath{stroke}%
\end{pgfscope}%
\begin{pgfscope}%
\pgfsetrectcap%
\pgfsetmiterjoin%
\pgfsetlinewidth{0.803000pt}%
\definecolor{currentstroke}{rgb}{0.000000,0.000000,0.000000}%
\pgfsetstrokecolor{currentstroke}%
\pgfsetdash{}{0pt}%
\pgfpathmoveto{\pgfqpoint{5.330808in}{2.335562in}}%
\pgfpathlineto{\pgfqpoint{5.951323in}{2.335562in}}%
\pgfusepath{stroke}%
\end{pgfscope}%
\begin{pgfscope}%
\definecolor{textcolor}{rgb}{0.000000,0.000000,0.000000}%
\pgfsetstrokecolor{textcolor}%
\pgfsetfillcolor{textcolor}%
\pgftext[x=5.641066in,y=2.418896in,,base]{\color{textcolor}\rmfamily\fontsize{9.600000}{11.520000}\selectfont Filter 1}%
\end{pgfscope}%
\begin{pgfscope}%
\pgfsetbuttcap%
\pgfsetmiterjoin%
\definecolor{currentfill}{rgb}{1.000000,1.000000,1.000000}%
\pgfsetfillcolor{currentfill}%
\pgfsetlinewidth{0.000000pt}%
\definecolor{currentstroke}{rgb}{0.000000,0.000000,0.000000}%
\pgfsetstrokecolor{currentstroke}%
\pgfsetstrokeopacity{0.000000}%
\pgfsetdash{}{0pt}%
\pgfpathmoveto{\pgfqpoint{4.648242in}{0.489808in}}%
\pgfpathlineto{\pgfqpoint{5.268757in}{0.489808in}}%
\pgfpathlineto{\pgfqpoint{5.268757in}{1.233062in}}%
\pgfpathlineto{\pgfqpoint{4.648242in}{1.233062in}}%
\pgfpathclose%
\pgfusepath{fill}%
\end{pgfscope}%
\begin{pgfscope}%
\pgfpathrectangle{\pgfqpoint{4.648242in}{0.489808in}}{\pgfqpoint{0.620515in}{0.743254in}}%
\pgfusepath{clip}%
\pgfsys@transformshift{4.648242in}{0.489808in}%
\pgftext[left,bottom]{\pgfimage[interpolate=true,width=0.624000in,height=0.744000in]{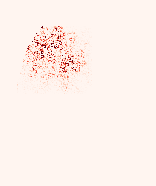}}%
\end{pgfscope}%
\begin{pgfscope}%
\pgfsetrectcap%
\pgfsetmiterjoin%
\pgfsetlinewidth{0.803000pt}%
\definecolor{currentstroke}{rgb}{0.000000,0.000000,0.000000}%
\pgfsetstrokecolor{currentstroke}%
\pgfsetdash{}{0pt}%
\pgfpathmoveto{\pgfqpoint{4.648242in}{0.489808in}}%
\pgfpathlineto{\pgfqpoint{4.648242in}{1.233062in}}%
\pgfusepath{stroke}%
\end{pgfscope}%
\begin{pgfscope}%
\pgfsetrectcap%
\pgfsetmiterjoin%
\pgfsetlinewidth{0.803000pt}%
\definecolor{currentstroke}{rgb}{0.000000,0.000000,0.000000}%
\pgfsetstrokecolor{currentstroke}%
\pgfsetdash{}{0pt}%
\pgfpathmoveto{\pgfqpoint{5.268757in}{0.489808in}}%
\pgfpathlineto{\pgfqpoint{5.268757in}{1.233062in}}%
\pgfusepath{stroke}%
\end{pgfscope}%
\begin{pgfscope}%
\pgfsetrectcap%
\pgfsetmiterjoin%
\pgfsetlinewidth{0.803000pt}%
\definecolor{currentstroke}{rgb}{0.000000,0.000000,0.000000}%
\pgfsetstrokecolor{currentstroke}%
\pgfsetdash{}{0pt}%
\pgfpathmoveto{\pgfqpoint{4.648242in}{0.489808in}}%
\pgfpathlineto{\pgfqpoint{5.268757in}{0.489808in}}%
\pgfusepath{stroke}%
\end{pgfscope}%
\begin{pgfscope}%
\pgfsetrectcap%
\pgfsetmiterjoin%
\pgfsetlinewidth{0.803000pt}%
\definecolor{currentstroke}{rgb}{0.000000,0.000000,0.000000}%
\pgfsetstrokecolor{currentstroke}%
\pgfsetdash{}{0pt}%
\pgfpathmoveto{\pgfqpoint{4.648242in}{1.233062in}}%
\pgfpathlineto{\pgfqpoint{5.268757in}{1.233062in}}%
\pgfusepath{stroke}%
\end{pgfscope}%
\begin{pgfscope}%
\definecolor{textcolor}{rgb}{0.000000,0.000000,0.000000}%
\pgfsetstrokecolor{textcolor}%
\pgfsetfillcolor{textcolor}%
\pgftext[x=4.958499in,y=1.316396in,,base]{\color{textcolor}\rmfamily\fontsize{9.600000}{11.520000}\selectfont Filter 2}%
\end{pgfscope}%
\begin{pgfscope}%
\pgfsetbuttcap%
\pgfsetmiterjoin%
\definecolor{currentfill}{rgb}{1.000000,1.000000,1.000000}%
\pgfsetfillcolor{currentfill}%
\pgfsetlinewidth{0.000000pt}%
\definecolor{currentstroke}{rgb}{0.000000,0.000000,0.000000}%
\pgfsetstrokecolor{currentstroke}%
\pgfsetstrokeopacity{0.000000}%
\pgfsetdash{}{0pt}%
\pgfpathmoveto{\pgfqpoint{5.330808in}{0.489808in}}%
\pgfpathlineto{\pgfqpoint{5.951323in}{0.489808in}}%
\pgfpathlineto{\pgfqpoint{5.951323in}{1.233062in}}%
\pgfpathlineto{\pgfqpoint{5.330808in}{1.233062in}}%
\pgfpathclose%
\pgfusepath{fill}%
\end{pgfscope}%
\begin{pgfscope}%
\pgfpathrectangle{\pgfqpoint{5.330808in}{0.489808in}}{\pgfqpoint{0.620515in}{0.743254in}}%
\pgfusepath{clip}%
\pgfsys@transformshift{5.330808in}{0.489808in}%
\pgftext[left,bottom]{\pgfimage[interpolate=true,width=0.624000in,height=0.744000in]{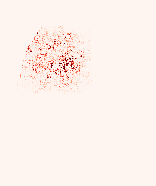}}%
\end{pgfscope}%
\begin{pgfscope}%
\pgfsetrectcap%
\pgfsetmiterjoin%
\pgfsetlinewidth{0.803000pt}%
\definecolor{currentstroke}{rgb}{0.000000,0.000000,0.000000}%
\pgfsetstrokecolor{currentstroke}%
\pgfsetdash{}{0pt}%
\pgfpathmoveto{\pgfqpoint{5.330808in}{0.489808in}}%
\pgfpathlineto{\pgfqpoint{5.330808in}{1.233062in}}%
\pgfusepath{stroke}%
\end{pgfscope}%
\begin{pgfscope}%
\pgfsetrectcap%
\pgfsetmiterjoin%
\pgfsetlinewidth{0.803000pt}%
\definecolor{currentstroke}{rgb}{0.000000,0.000000,0.000000}%
\pgfsetstrokecolor{currentstroke}%
\pgfsetdash{}{0pt}%
\pgfpathmoveto{\pgfqpoint{5.951323in}{0.489808in}}%
\pgfpathlineto{\pgfqpoint{5.951323in}{1.233062in}}%
\pgfusepath{stroke}%
\end{pgfscope}%
\begin{pgfscope}%
\pgfsetrectcap%
\pgfsetmiterjoin%
\pgfsetlinewidth{0.803000pt}%
\definecolor{currentstroke}{rgb}{0.000000,0.000000,0.000000}%
\pgfsetstrokecolor{currentstroke}%
\pgfsetdash{}{0pt}%
\pgfpathmoveto{\pgfqpoint{5.330808in}{0.489808in}}%
\pgfpathlineto{\pgfqpoint{5.951323in}{0.489808in}}%
\pgfusepath{stroke}%
\end{pgfscope}%
\begin{pgfscope}%
\pgfsetrectcap%
\pgfsetmiterjoin%
\pgfsetlinewidth{0.803000pt}%
\definecolor{currentstroke}{rgb}{0.000000,0.000000,0.000000}%
\pgfsetstrokecolor{currentstroke}%
\pgfsetdash{}{0pt}%
\pgfpathmoveto{\pgfqpoint{5.330808in}{1.233062in}}%
\pgfpathlineto{\pgfqpoint{5.951323in}{1.233062in}}%
\pgfusepath{stroke}%
\end{pgfscope}%
\begin{pgfscope}%
\definecolor{textcolor}{rgb}{0.000000,0.000000,0.000000}%
\pgfsetstrokecolor{textcolor}%
\pgfsetfillcolor{textcolor}%
\pgftext[x=5.641066in,y=1.316396in,,base]{\color{textcolor}\rmfamily\fontsize{9.600000}{11.520000}\selectfont Filter 4}%
\end{pgfscope}%
\begin{pgfscope}%
\definecolor{textcolor}{rgb}{0.000000,0.000000,0.000000}%
\pgfsetstrokecolor{textcolor}%
\pgfsetfillcolor{textcolor}%
\pgftext[x=3.059323in,y=0.235185in,,top]{\color{textcolor}\rmfamily\fontsize{11.000000}{13.200000}\bfseries\selectfont LRP heatmaps of PIF model}%
\end{pgfscope}%
\end{pgfpicture}%
\makeatother%
\endgroup%

%% file: figures/LRP_heatmaps_experiment_conv4+score_UKB.pgf
\begingroup%
\makeatletter%
\begin{pgfpicture}%
\pgfpathrectangle{\pgfpointorigin}{\pgfqpoint{4.337073in}{3.322969in}}%
\pgfusepath{use as bounding box, clip}%
\begin{pgfscope}%
\pgfsetbuttcap%
\pgfsetmiterjoin%
\definecolor{currentfill}{rgb}{1.000000,1.000000,1.000000}%
\pgfsetfillcolor{currentfill}%
\pgfsetlinewidth{0.000000pt}%
\definecolor{currentstroke}{rgb}{1.000000,1.000000,1.000000}%
\pgfsetstrokecolor{currentstroke}%
\pgfsetdash{}{0pt}%
\pgfpathmoveto{\pgfqpoint{0.000000in}{-0.000000in}}%
\pgfpathlineto{\pgfqpoint{4.337073in}{-0.000000in}}%
\pgfpathlineto{\pgfqpoint{4.337073in}{3.322969in}}%
\pgfpathlineto{\pgfqpoint{0.000000in}{3.322969in}}%
\pgfpathclose%
\pgfusepath{fill}%
\end{pgfscope}%
\begin{pgfscope}%
\definecolor{textcolor}{rgb}{0.000000,0.000000,0.000000}%
\pgfsetstrokecolor{textcolor}%
\pgfsetfillcolor{textcolor}%
\pgftext[x=1.270692in,y=3.126518in,,base]{\color{textcolor}\rmfamily\fontsize{9.600000}{11.520000}\bfseries\selectfont Layer Conv4}%
\end{pgfscope}%
\begin{pgfscope}%
\pgfsetbuttcap%
\pgfsetmiterjoin%
\definecolor{currentfill}{rgb}{1.000000,1.000000,1.000000}%
\pgfsetfillcolor{currentfill}%
\pgfsetlinewidth{0.000000pt}%
\definecolor{currentstroke}{rgb}{0.000000,0.000000,0.000000}%
\pgfsetstrokecolor{currentstroke}%
\pgfsetstrokeopacity{0.000000}%
\pgfsetdash{}{0pt}%
\pgfpathmoveto{\pgfqpoint{0.348073in}{1.869797in}}%
\pgfpathlineto{\pgfqpoint{1.226758in}{1.869797in}}%
\pgfpathlineto{\pgfqpoint{1.226758in}{2.922288in}}%
\pgfpathlineto{\pgfqpoint{0.348073in}{2.922288in}}%
\pgfpathclose%
\pgfusepath{fill}%
\end{pgfscope}%
\begin{pgfscope}%
\pgfpathrectangle{\pgfqpoint{0.348073in}{1.869797in}}{\pgfqpoint{0.878685in}{1.052491in}}%
\pgfusepath{clip}%
\pgfsys@transformshift{0.348073in}{1.869797in}%
\pgftext[left,bottom]{\pgfimage[interpolate=true,width=0.880000in,height=1.056000in]{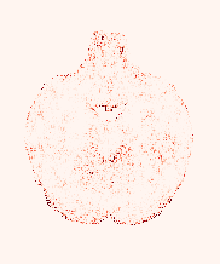}}%
\end{pgfscope}%
\begin{pgfscope}%
\pgfsetrectcap%
\pgfsetmiterjoin%
\pgfsetlinewidth{0.803000pt}%
\definecolor{currentstroke}{rgb}{0.000000,0.000000,0.000000}%
\pgfsetstrokecolor{currentstroke}%
\pgfsetdash{}{0pt}%
\pgfpathmoveto{\pgfqpoint{0.348073in}{1.869797in}}%
\pgfpathlineto{\pgfqpoint{0.348073in}{2.922288in}}%
\pgfusepath{stroke}%
\end{pgfscope}%
\begin{pgfscope}%
\pgfsetrectcap%
\pgfsetmiterjoin%
\pgfsetlinewidth{0.803000pt}%
\definecolor{currentstroke}{rgb}{0.000000,0.000000,0.000000}%
\pgfsetstrokecolor{currentstroke}%
\pgfsetdash{}{0pt}%
\pgfpathmoveto{\pgfqpoint{1.226758in}{1.869797in}}%
\pgfpathlineto{\pgfqpoint{1.226758in}{2.922288in}}%
\pgfusepath{stroke}%
\end{pgfscope}%
\begin{pgfscope}%
\pgfsetrectcap%
\pgfsetmiterjoin%
\pgfsetlinewidth{0.803000pt}%
\definecolor{currentstroke}{rgb}{0.000000,0.000000,0.000000}%
\pgfsetstrokecolor{currentstroke}%
\pgfsetdash{}{0pt}%
\pgfpathmoveto{\pgfqpoint{0.348073in}{1.869797in}}%
\pgfpathlineto{\pgfqpoint{1.226758in}{1.869797in}}%
\pgfusepath{stroke}%
\end{pgfscope}%
\begin{pgfscope}%
\pgfsetrectcap%
\pgfsetmiterjoin%
\pgfsetlinewidth{0.803000pt}%
\definecolor{currentstroke}{rgb}{0.000000,0.000000,0.000000}%
\pgfsetstrokecolor{currentstroke}%
\pgfsetdash{}{0pt}%
\pgfpathmoveto{\pgfqpoint{0.348073in}{2.922288in}}%
\pgfpathlineto{\pgfqpoint{1.226758in}{2.922288in}}%
\pgfusepath{stroke}%
\end{pgfscope}%
\begin{pgfscope}%
\definecolor{textcolor}{rgb}{0.000000,0.000000,0.000000}%
\pgfsetstrokecolor{textcolor}%
\pgfsetfillcolor{textcolor}%
\pgftext[x=0.787416in,y=3.005621in,,base]{\color{textcolor}\rmfamily\fontsize{9.600000}{11.520000}\selectfont Filter 0}%
\end{pgfscope}%
\begin{pgfscope}%
\pgfsetbuttcap%
\pgfsetmiterjoin%
\definecolor{currentfill}{rgb}{1.000000,1.000000,1.000000}%
\pgfsetfillcolor{currentfill}%
\pgfsetlinewidth{0.000000pt}%
\definecolor{currentstroke}{rgb}{0.000000,0.000000,0.000000}%
\pgfsetstrokecolor{currentstroke}%
\pgfsetstrokeopacity{0.000000}%
\pgfsetdash{}{0pt}%
\pgfpathmoveto{\pgfqpoint{1.314626in}{1.869797in}}%
\pgfpathlineto{\pgfqpoint{2.193311in}{1.869797in}}%
\pgfpathlineto{\pgfqpoint{2.193311in}{2.922288in}}%
\pgfpathlineto{\pgfqpoint{1.314626in}{2.922288in}}%
\pgfpathclose%
\pgfusepath{fill}%
\end{pgfscope}%
\begin{pgfscope}%
\pgfpathrectangle{\pgfqpoint{1.314626in}{1.869797in}}{\pgfqpoint{0.878685in}{1.052491in}}%
\pgfusepath{clip}%
\pgfsys@transformshift{1.314626in}{1.869797in}%
\pgftext[left,bottom]{\pgfimage[interpolate=true,width=0.880000in,height=1.056000in]{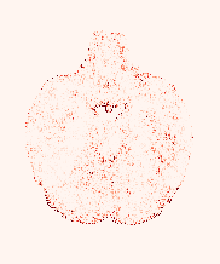}}%
\end{pgfscope}%
\begin{pgfscope}%
\pgfsetrectcap%
\pgfsetmiterjoin%
\pgfsetlinewidth{0.803000pt}%
\definecolor{currentstroke}{rgb}{0.000000,0.000000,0.000000}%
\pgfsetstrokecolor{currentstroke}%
\pgfsetdash{}{0pt}%
\pgfpathmoveto{\pgfqpoint{1.314626in}{1.869797in}}%
\pgfpathlineto{\pgfqpoint{1.314626in}{2.922288in}}%
\pgfusepath{stroke}%
\end{pgfscope}%
\begin{pgfscope}%
\pgfsetrectcap%
\pgfsetmiterjoin%
\pgfsetlinewidth{0.803000pt}%
\definecolor{currentstroke}{rgb}{0.000000,0.000000,0.000000}%
\pgfsetstrokecolor{currentstroke}%
\pgfsetdash{}{0pt}%
\pgfpathmoveto{\pgfqpoint{2.193311in}{1.869797in}}%
\pgfpathlineto{\pgfqpoint{2.193311in}{2.922288in}}%
\pgfusepath{stroke}%
\end{pgfscope}%
\begin{pgfscope}%
\pgfsetrectcap%
\pgfsetmiterjoin%
\pgfsetlinewidth{0.803000pt}%
\definecolor{currentstroke}{rgb}{0.000000,0.000000,0.000000}%
\pgfsetstrokecolor{currentstroke}%
\pgfsetdash{}{0pt}%
\pgfpathmoveto{\pgfqpoint{1.314626in}{1.869797in}}%
\pgfpathlineto{\pgfqpoint{2.193311in}{1.869797in}}%
\pgfusepath{stroke}%
\end{pgfscope}%
\begin{pgfscope}%
\pgfsetrectcap%
\pgfsetmiterjoin%
\pgfsetlinewidth{0.803000pt}%
\definecolor{currentstroke}{rgb}{0.000000,0.000000,0.000000}%
\pgfsetstrokecolor{currentstroke}%
\pgfsetdash{}{0pt}%
\pgfpathmoveto{\pgfqpoint{1.314626in}{2.922288in}}%
\pgfpathlineto{\pgfqpoint{2.193311in}{2.922288in}}%
\pgfusepath{stroke}%
\end{pgfscope}%
\begin{pgfscope}%
\definecolor{textcolor}{rgb}{0.000000,0.000000,0.000000}%
\pgfsetstrokecolor{textcolor}%
\pgfsetfillcolor{textcolor}%
\pgftext[x=1.753969in,y=3.005621in,,base]{\color{textcolor}\rmfamily\fontsize{9.600000}{11.520000}\selectfont Filter 4}%
\end{pgfscope}%
\begin{pgfscope}%
\pgfsetbuttcap%
\pgfsetmiterjoin%
\definecolor{currentfill}{rgb}{1.000000,1.000000,1.000000}%
\pgfsetfillcolor{currentfill}%
\pgfsetlinewidth{0.000000pt}%
\definecolor{currentstroke}{rgb}{0.000000,0.000000,0.000000}%
\pgfsetstrokecolor{currentstroke}%
\pgfsetstrokeopacity{0.000000}%
\pgfsetdash{}{0pt}%
\pgfpathmoveto{\pgfqpoint{0.348073in}{0.446083in}}%
\pgfpathlineto{\pgfqpoint{1.226758in}{0.446083in}}%
\pgfpathlineto{\pgfqpoint{1.226758in}{1.498573in}}%
\pgfpathlineto{\pgfqpoint{0.348073in}{1.498573in}}%
\pgfpathclose%
\pgfusepath{fill}%
\end{pgfscope}%
\begin{pgfscope}%
\pgfpathrectangle{\pgfqpoint{0.348073in}{0.446083in}}{\pgfqpoint{0.878685in}{1.052491in}}%
\pgfusepath{clip}%
\pgfsys@transformshift{0.348073in}{0.446083in}%
\pgftext[left,bottom]{\pgfimage[interpolate=true,width=0.880000in,height=1.056000in]{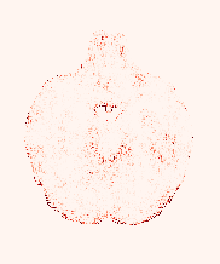}}%
\end{pgfscope}%
\begin{pgfscope}%
\pgfsetrectcap%
\pgfsetmiterjoin%
\pgfsetlinewidth{0.803000pt}%
\definecolor{currentstroke}{rgb}{0.000000,0.000000,0.000000}%
\pgfsetstrokecolor{currentstroke}%
\pgfsetdash{}{0pt}%
\pgfpathmoveto{\pgfqpoint{0.348073in}{0.446083in}}%
\pgfpathlineto{\pgfqpoint{0.348073in}{1.498573in}}%
\pgfusepath{stroke}%
\end{pgfscope}%
\begin{pgfscope}%
\pgfsetrectcap%
\pgfsetmiterjoin%
\pgfsetlinewidth{0.803000pt}%
\definecolor{currentstroke}{rgb}{0.000000,0.000000,0.000000}%
\pgfsetstrokecolor{currentstroke}%
\pgfsetdash{}{0pt}%
\pgfpathmoveto{\pgfqpoint{1.226758in}{0.446083in}}%
\pgfpathlineto{\pgfqpoint{1.226758in}{1.498573in}}%
\pgfusepath{stroke}%
\end{pgfscope}%
\begin{pgfscope}%
\pgfsetrectcap%
\pgfsetmiterjoin%
\pgfsetlinewidth{0.803000pt}%
\definecolor{currentstroke}{rgb}{0.000000,0.000000,0.000000}%
\pgfsetstrokecolor{currentstroke}%
\pgfsetdash{}{0pt}%
\pgfpathmoveto{\pgfqpoint{0.348073in}{0.446083in}}%
\pgfpathlineto{\pgfqpoint{1.226758in}{0.446083in}}%
\pgfusepath{stroke}%
\end{pgfscope}%
\begin{pgfscope}%
\pgfsetrectcap%
\pgfsetmiterjoin%
\pgfsetlinewidth{0.803000pt}%
\definecolor{currentstroke}{rgb}{0.000000,0.000000,0.000000}%
\pgfsetstrokecolor{currentstroke}%
\pgfsetdash{}{0pt}%
\pgfpathmoveto{\pgfqpoint{0.348073in}{1.498573in}}%
\pgfpathlineto{\pgfqpoint{1.226758in}{1.498573in}}%
\pgfusepath{stroke}%
\end{pgfscope}%
\begin{pgfscope}%
\definecolor{textcolor}{rgb}{0.000000,0.000000,0.000000}%
\pgfsetstrokecolor{textcolor}%
\pgfsetfillcolor{textcolor}%
\pgftext[x=0.787416in,y=1.581907in,,base]{\color{textcolor}\rmfamily\fontsize{9.600000}{11.520000}\selectfont Filter 15}%
\end{pgfscope}%
\begin{pgfscope}%
\pgfsetbuttcap%
\pgfsetmiterjoin%
\definecolor{currentfill}{rgb}{1.000000,1.000000,1.000000}%
\pgfsetfillcolor{currentfill}%
\pgfsetlinewidth{0.000000pt}%
\definecolor{currentstroke}{rgb}{0.000000,0.000000,0.000000}%
\pgfsetstrokecolor{currentstroke}%
\pgfsetstrokeopacity{0.000000}%
\pgfsetdash{}{0pt}%
\pgfpathmoveto{\pgfqpoint{1.314626in}{0.446083in}}%
\pgfpathlineto{\pgfqpoint{2.193311in}{0.446083in}}%
\pgfpathlineto{\pgfqpoint{2.193311in}{1.498573in}}%
\pgfpathlineto{\pgfqpoint{1.314626in}{1.498573in}}%
\pgfpathclose%
\pgfusepath{fill}%
\end{pgfscope}%
\begin{pgfscope}%
\pgfpathrectangle{\pgfqpoint{1.314626in}{0.446083in}}{\pgfqpoint{0.878685in}{1.052491in}}%
\pgfusepath{clip}%
\pgfsys@transformshift{1.314626in}{0.446083in}%
\pgftext[left,bottom]{\pgfimage[interpolate=true,width=0.880000in,height=1.056000in]{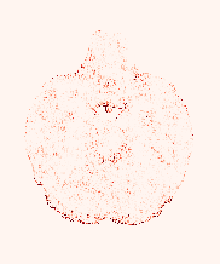}}%
\end{pgfscope}%
\begin{pgfscope}%
\pgfsetrectcap%
\pgfsetmiterjoin%
\pgfsetlinewidth{0.803000pt}%
\definecolor{currentstroke}{rgb}{0.000000,0.000000,0.000000}%
\pgfsetstrokecolor{currentstroke}%
\pgfsetdash{}{0pt}%
\pgfpathmoveto{\pgfqpoint{1.314626in}{0.446083in}}%
\pgfpathlineto{\pgfqpoint{1.314626in}{1.498573in}}%
\pgfusepath{stroke}%
\end{pgfscope}%
\begin{pgfscope}%
\pgfsetrectcap%
\pgfsetmiterjoin%
\pgfsetlinewidth{0.803000pt}%
\definecolor{currentstroke}{rgb}{0.000000,0.000000,0.000000}%
\pgfsetstrokecolor{currentstroke}%
\pgfsetdash{}{0pt}%
\pgfpathmoveto{\pgfqpoint{2.193311in}{0.446083in}}%
\pgfpathlineto{\pgfqpoint{2.193311in}{1.498573in}}%
\pgfusepath{stroke}%
\end{pgfscope}%
\begin{pgfscope}%
\pgfsetrectcap%
\pgfsetmiterjoin%
\pgfsetlinewidth{0.803000pt}%
\definecolor{currentstroke}{rgb}{0.000000,0.000000,0.000000}%
\pgfsetstrokecolor{currentstroke}%
\pgfsetdash{}{0pt}%
\pgfpathmoveto{\pgfqpoint{1.314626in}{0.446083in}}%
\pgfpathlineto{\pgfqpoint{2.193311in}{0.446083in}}%
\pgfusepath{stroke}%
\end{pgfscope}%
\begin{pgfscope}%
\pgfsetrectcap%
\pgfsetmiterjoin%
\pgfsetlinewidth{0.803000pt}%
\definecolor{currentstroke}{rgb}{0.000000,0.000000,0.000000}%
\pgfsetstrokecolor{currentstroke}%
\pgfsetdash{}{0pt}%
\pgfpathmoveto{\pgfqpoint{1.314626in}{1.498573in}}%
\pgfpathlineto{\pgfqpoint{2.193311in}{1.498573in}}%
\pgfusepath{stroke}%
\end{pgfscope}%
\begin{pgfscope}%
\definecolor{textcolor}{rgb}{0.000000,0.000000,0.000000}%
\pgfsetstrokecolor{textcolor}%
\pgfsetfillcolor{textcolor}%
\pgftext[x=1.753969in,y=1.581907in,,base]{\color{textcolor}\rmfamily\fontsize{9.600000}{11.520000}\selectfont Filter 60}%
\end{pgfscope}%
\begin{pgfscope}%
\pgfsetbuttcap%
\pgfsetmiterjoin%
\definecolor{currentfill}{rgb}{1.000000,1.000000,1.000000}%
\pgfsetfillcolor{currentfill}%
\pgfsetlinewidth{0.000000pt}%
\definecolor{currentstroke}{rgb}{0.000000,0.000000,0.000000}%
\pgfsetstrokecolor{currentstroke}%
\pgfsetstrokeopacity{0.000000}%
\pgfsetdash{}{0pt}%
\pgfpathmoveto{\pgfqpoint{2.377835in}{0.579070in}}%
\pgfpathlineto{\pgfqpoint{4.223073in}{0.579070in}}%
\pgfpathlineto{\pgfqpoint{4.223073in}{2.789300in}}%
\pgfpathlineto{\pgfqpoint{2.377835in}{2.789300in}}%
\pgfpathclose%
\pgfusepath{fill}%
\end{pgfscope}%
\begin{pgfscope}%
\pgfpathrectangle{\pgfqpoint{2.377835in}{0.579070in}}{\pgfqpoint{1.845238in}{2.210230in}}%
\pgfusepath{clip}%
\pgfsys@transformshift{2.377835in}{0.579070in}%
\pgftext[left,bottom]{\pgfimage[interpolate=true,width=1.848000in,height=2.212000in]{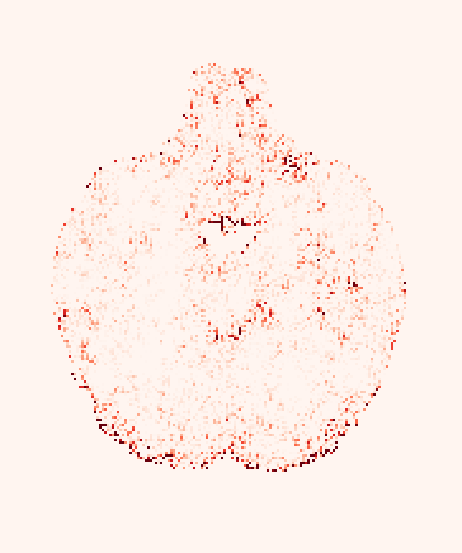}}%
\end{pgfscope}%
\begin{pgfscope}%
\pgfsetrectcap%
\pgfsetmiterjoin%
\pgfsetlinewidth{0.803000pt}%
\definecolor{currentstroke}{rgb}{0.000000,0.000000,0.000000}%
\pgfsetstrokecolor{currentstroke}%
\pgfsetdash{}{0pt}%
\pgfpathmoveto{\pgfqpoint{2.377835in}{0.579070in}}%
\pgfpathlineto{\pgfqpoint{2.377835in}{2.789300in}}%
\pgfusepath{stroke}%
\end{pgfscope}%
\begin{pgfscope}%
\pgfsetrectcap%
\pgfsetmiterjoin%
\pgfsetlinewidth{0.803000pt}%
\definecolor{currentstroke}{rgb}{0.000000,0.000000,0.000000}%
\pgfsetstrokecolor{currentstroke}%
\pgfsetdash{}{0pt}%
\pgfpathmoveto{\pgfqpoint{4.223073in}{0.579070in}}%
\pgfpathlineto{\pgfqpoint{4.223073in}{2.789300in}}%
\pgfusepath{stroke}%
\end{pgfscope}%
\begin{pgfscope}%
\pgfsetrectcap%
\pgfsetmiterjoin%
\pgfsetlinewidth{0.803000pt}%
\definecolor{currentstroke}{rgb}{0.000000,0.000000,0.000000}%
\pgfsetstrokecolor{currentstroke}%
\pgfsetdash{}{0pt}%
\pgfpathmoveto{\pgfqpoint{2.377835in}{0.579070in}}%
\pgfpathlineto{\pgfqpoint{4.223073in}{0.579070in}}%
\pgfusepath{stroke}%
\end{pgfscope}%
\begin{pgfscope}%
\pgfsetrectcap%
\pgfsetmiterjoin%
\pgfsetlinewidth{0.803000pt}%
\definecolor{currentstroke}{rgb}{0.000000,0.000000,0.000000}%
\pgfsetstrokecolor{currentstroke}%
\pgfsetdash{}{0pt}%
\pgfpathmoveto{\pgfqpoint{2.377835in}{2.789300in}}%
\pgfpathlineto{\pgfqpoint{4.223073in}{2.789300in}}%
\pgfusepath{stroke}%
\end{pgfscope}%
\begin{pgfscope}%
\definecolor{textcolor}{rgb}{0.000000,0.000000,0.000000}%
\pgfsetstrokecolor{textcolor}%
\pgfsetfillcolor{textcolor}%
\pgftext[x=3.300454in,y=2.872634in,,base]{\color{textcolor}\rmfamily\fontsize{9.600000}{11.520000}\bfseries\selectfont Score/Final Output}%
\end{pgfscope}%
\begin{pgfscope}%
\definecolor{textcolor}{rgb}{0.000000,0.000000,0.000000}%
\pgfsetstrokecolor{textcolor}%
\pgfsetfillcolor{textcolor}%
\pgftext[x=2.223073in,y=0.235185in,,top]{\color{textcolor}\rmfamily\fontsize{11.000000}{13.200000}\bfseries\selectfont PIF model LRP heatmaps}%
\end{pgfscope}%
\end{pgfpicture}%
\makeatother%
\endgroup%